\documentclass{article}
\usepackage{oct_preprint,times}
\usepackage[T1]{fontenc}
\usepackage{amsmath,amssymb,graphicx,booktabs,tabularx,array,multirow}
\usepackage{xcolor,placeins}
\usepackage{hyperref,url}

\usepackage{siunitx}
\DeclareSIUnit\inch{in.}
\DeclareSIUnit[quantity-product = ]\factor{\times}
\DeclareSIUnit\sample{S}
\DeclareSIUnit\pixels{px}
\DeclareSIUnit\voxels{vx}

\usepackage{cleveref}

\hypersetup{hidelinks,
    pdftitle={Robotic Optical Coherence Tomography-Guided Geometric Modeling of Soft Tissue: 3D Representations for Data-Driven Closed-Loop Shape Control},
    pdfauthor={Chaoyu Zhang, Guangshen Ma, Mark Draelos}
}

\title{Robotic Optical Coherence Tomography-Guided Geometric Modeling of Soft Tissue: 3D Representations for Data-Driven Closed-Loop Shape Control}

\author{
Chaoyu Zhang$^{*}$ \qquad
Guangshen Ma, PhD$^{~*,\dagger}$ \qquad
Mark Draelos, MD, PhD \\
\normalfont  $^{*}$Equal contribution. \quad $^{\dagger}$Project lead. \\ 
\normalfont University of Michigan, Ann Arbor \\
}
\begin{document}
\maketitle

\begin{abstract}

In medical robotics, data-driven soft-tissue shape control requires deforming tissue toward desired morphologies through sequential physical interactions.
This challenge is particularly important in microscale manipulation, where localized deformation is difficult to observe, represent, and control, while physics-based models can be computationally expensive and difficult to adapt to realistic environments.
Optical coherence tomography (OCT) provides high-resolution volumetric perception of tissue deformation, but target-conditioned control requires effective representations of both observed and desired tissue states.
In this work, we introduce an OCT-guided data-driven framework for tissue shape control, learned from user-guided demonstrations and executed with closed-loop OCT feedback.
We study full-volume, tissue-masked sparse-volume, and voxelized point-cloud representations, with or without residual conditioning, using deterministic (Explicit) and generative diffusion (Diffusion) policies.
Across 12 configurations and 576 phantom rollouts on 3-DoF depth and 6-DoF directional tasks, OCT-derived 3D representations support both Explicit and Diffusion policies, with performance depending on representation and residual conditioning.
Residual conditioning reduces directional Shape RMSE from 0.748 to 0.555 mm for full-volume Diffusion, while voxelized point-cloud Explicit with residuals achieves 0.534 mm for depth and 0.443 mm for directional control, with comparable accuracy at intermediate unseen targets and consistent performance across deployments.
We further evaluate the framework in 60 \textit{ex vivo} porcine rollouts, extending to biological tissue with final geometric errors comparable to human demonstrations.
These results establish volumetric OCT as current and goal representations for learned closed-loop tissue shape control across deterministic and generative action models.

\end{abstract}

\section{Introduction}
\label{sec:intro}

In medical robotics, microscale tissue manipulation requires precise control of deformable tissue through physical interaction, where tissue configuration can directly affect task execution~\citep{thach2024defgoalnet,wang2023goal}.
For example, robotic palpation relies on controlled tool-tissue interaction to induce and regulate local deformation toward a desired morphological state.
The control objective is therefore not simply a tool pose, but a target tissue morphology achieved through sequential interactions and successive observation of tissue response.
Goal-conditioned deformable manipulation has demonstrated this principle using image contours and point clouds~\citep{navarro2018fourier,thach2022deformernet,thach2024defgoalnet}.
Extending this paradigm to volumetric medical imaging is particularly relevant to microscale manipulation, where 3D image volumes preserve spatial and structural information~\citep{zhou2026volumedp} and allow desired tissue morphologies to be represented directly in the imaging space using both intensity-based and geometric representations.

\begin{figure}[h]
\centering
\includegraphics[width=\linewidth]{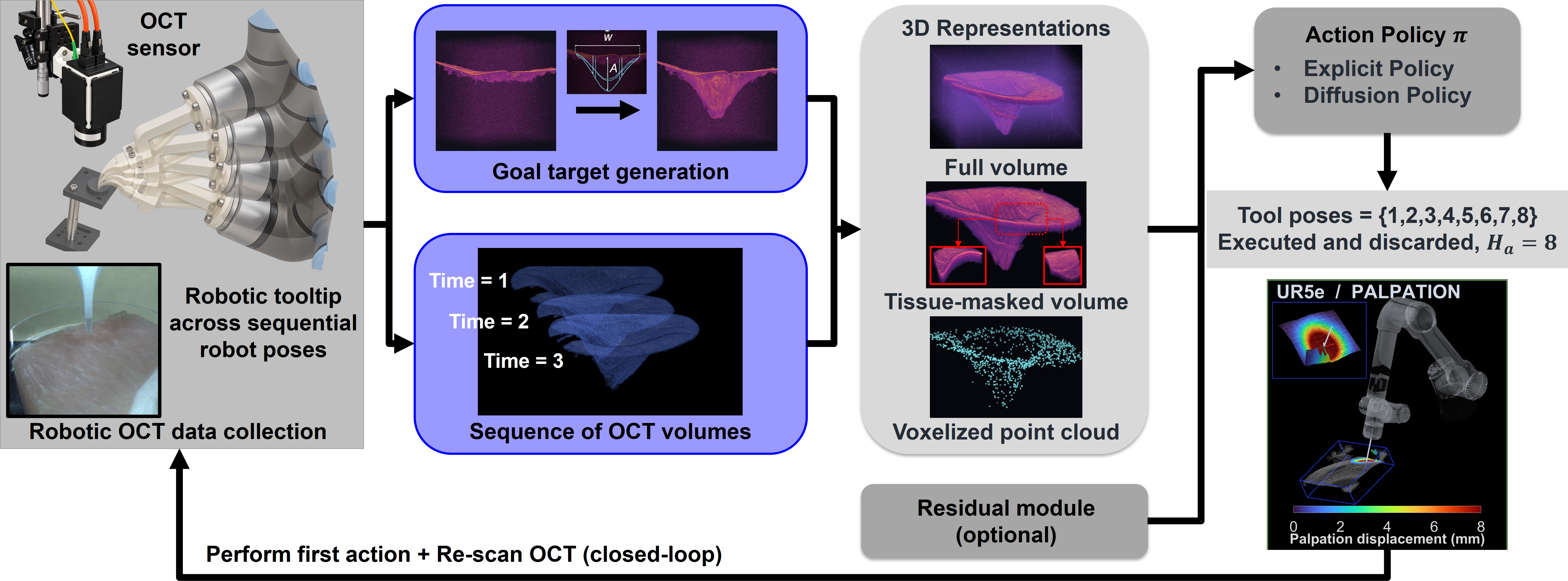}
\caption{
\textbf{Closed-loop OCT-guided tissue shape control.}
The OCT captures volumetric tissue deformation induced by the robotic tooltip, while parameterized target generation constructs desired morphologies from predeformed OCT states.
Current and target OCT volumes are encoded as full volumes, tissue-masked volumes, or voxelized point clouds, with optional current--target residual conditioning.
These representations condition an Explicit or Diffusion policy to predict tooltip motions, with the first action executed before an OCT scan updates the state for closed-loop prediction.
}
\label{fig:closeloopsys}
\end{figure}

At the microscale, morphology-based control requires high-resolution volumetric observations that capture both surface geometry and depth-dependent tissue structure.
Optical coherence tomography (OCT) provides such observations, resolving local surface morphology and subsurface structure within a 3D imaging volume~\citep{huang1991optical, ma2025robotics}.
This volumetric perception has enabled robotic OCT systems for learned interaction, insertion control, and deformation-aware instrument navigation~\citep{keller2020rl,edwards2022dalk,arikan2025deformationaware}.
Based on these capabilities, we develop \textbf{\emph{OCT-guided soft-tissue shape control}}, a learning-based framework (Fig.~\ref{fig:closeloopsys}) that uses OCT feedback to guide tool motion toward a target tissue morphology represented in the same imaging space.
This closed-loop formulation enables the robot to adapt its actions to observed deformation, providing a foundation for image-guided tissue shaping in microscale interventions.

Realizing closed-loop OCT-guided shape control requires jointly addressing target construction, OCT representation, and action prediction. 
Prior work has used geometric and learned goal shapes for deformable-object manipulation~\citep{thach2022deformernet,thach2026diffdef}, while OCT-derived point clouds have been explored for registration and reconstruction~\citep{liu2024c2p,fang2025octmontaging}. 
Building upon these directions, we introduce an OCT-guided robotic framework that constructs parameterized target morphologies from measured tissue states and learns to deform tissue toward these targets from user demonstrations. 
We investigate full OCT volumes, tissue-masked intensity volumes, and voxelized point clouds for representing current and target states, and compare deterministic Explicit and Diffusion action models~\citep{chi2023diffusionpolicy,ze2024dp3}, with and without representation-specific current--target residual conditioning. 
These policies predict incremental 6-DoF tooltip motions, while successive OCT scans provide closed-loop feedback toward the desired tissue morphology.

In this work, we evaluate 12 policy configurations on physical robotic shape control tasks spanning depth and directional deformations, including intermediate targets unseen during training, and further demonstrate the framework on \textit{ex vivo} porcine tissue with varying surface geometries.
Our results demonstrate the feasibility of using volumetric OCT observations for closed-loop shape control with both Explicit and Diffusion policies, while voxelized point cloud representations provide accurate and repeatable control through compact geometric encoding.
In addition, current-to-target residual conditioning further improves performance across most configurations and strengthens target-dependent responses, particularly for full volume representations.
Overall, this work establishes a systematic framework for studying how volumetric and geometric OCT representations support learned closed-loop tissue shape control, connecting high-resolution medical imaging with deterministic Explicit and generative Diffusion action policies for deformable tissue manipulation.

\section{Related Work}
\label{sec:related}

\noindent
\textbf{OCT-guided robotic intervention:}
OCT has increasingly been integrated into robotic intervention through image-based visual servoing~\citep{duflot2018octservo}, learning-based control from demonstrations~\citep{keller2020rl}, and data-driven modeling of tool--tissue interactions~\citep{edwards2022dalk,ma2025robotics}.
For example, in retinal interventions, OCT-guided systems combine instrument localization, motion planning, and feedback control for autonomous navigation and targeted injection~\citep{zhang2024autonomousneedle,bian2026arise, liao2026automatic}, while deformation-aware approaches incorporate anatomical changes to adapt instrument positioning during tissue interaction~\citep{arikan2025deformationaware, draelos2019optical}.
These methods primarily use OCT to localize anatomy, guide instruments, or adapt trajectories relative to anatomical targets.
In contrast, we formulate the observed tissue morphology as the controlled state and a desired 3D tissue configuration as the goal, extending OCT-guided robotic intervention toward target-conditioned volumetric tissue shape control.

\noindent
\textbf{OCT representations for geometry and interaction:}
Learning from OCT has explored both volumetric appearance and explicit tissue geometry.
Three-dimensional convolutional networks have been used to estimate interaction forces directly from OCT volumes~\citep{gessert2018force}, while spatiotemporal models capture tissue deformation across successive scans~\citep{gessert2020fourdimensional}.
Beyond volumetric intensity, geometric learning on segmented OCT point clouds has been compared with volumetric convolution for glaucoma diagnosis~\citep{thiery2023geometric}.
Point-cloud representations also support anatomical reconstruction: C2P-Net performs non-rigid registration of anatomical templates to partial OCT point clouds~\citep{liu2024c2p}, while point cloud-based volumetric montaging integrates robotic OCT acquisitions into an expanded field of view~\citep{fang2025octmontaging}.
For tissue interaction modeling, OCT-DeformNet uses OCT-derived surface geometry and robot tool motion to predict biological tissue deformation~\citep{ma2026octdeformnet}.
Though these studies demonstrate the utility of different OCT representations for perception, reconstruction, and deformation prediction, we investigate their role in closed-loop action learning by comparing volumetric intensity, tissue-masked intensity, and point geometry for target-conditioned physical tissue deformation.

\noindent
\textbf{Goal conditioning and action prediction for deformable manipulation:}
Robotic shape control requires both specifying desired configurations and translating current--goal relationships into effective manipulation actions.
Prior goal-conditioned shape servoing represents targets using image contours~\citep{navarro2018fourier} or partial-view point clouds~\citep{thach2022deformernet}.
Beyond explicitly specified goals, DefGoalNet learns context-dependent target shapes from demonstrations~\citep{thach2024defgoalnet}, while DiffDef models multimodal goal distributions via diffusion and passes generated goals to DeformerNet for action prediction~\citep{thach2026diffdef}.
Model-based approaches such as RoboCraft and DexDeform instead leverage learned dynamics or differentiable physics for deformable-object planning~\citep{shi2022robocraft,li2023dexdeform}.
In this work, we construct volumetric OCT targets by applying parameterized deformations to measured tissue states, enabling direct control over local target depth, width, and direction while preserving specimen-specific OCT structure.
These targets support both intensity-based and geometric representations, enabling systematic evaluation of how 3D goal representation affects closed-loop tissue shape control.

Given current and desired states, action prediction translates their relationship into goal-directed manipulation.
Deterministic policies directly regress actions, whereas Diffusion Policy generates visuomotor action sequences through conditional denoising~\citep{chi2023diffusionpolicy,wolf2025diffusion}.
Reuss et al.\ incorporate explicit goals into score-based action generation~\citep{reuss2023beso}, while DP3 combines compact point-cloud representations with diffusion-based action prediction~\citep{ze2024dp3}.
Movement Primitive Diffusion further integrates diffusion-based imitation learning with structured motion generation for deformable-object manipulation~\citep{scheikl2024mpd}.
Goal conditioning can explicitly encode current--goal discrepancies, as in DeformerNet's use of feature differences~\citep{thach2022deformernet}.
Motivated by these approaches, we compare Explicit and Diffusion policies with and without representation-specific current--target residual conditioning, examining how action prediction and goal conditioning interact with the information retained by different OCT representations.

\section{OCT-Guided Tissue Shape Control Framework}
\label{sec:method}

We formulate soft-tissue shape control as a goal-conditioned robot learning problem with closed-loop volumetric OCT feedback.
A policy learned from user-guided demonstrations predicts tooltip motions conditioned on current OCT observations and a target tissue morphology.
The framework integrates parameterized target generation (Appendix~\ref{sec:goals}), 3D OCT representations, goal-conditioning schemes, and Explicit and Diffusion policies within a unified closed-loop formulation.

\subsection{Problem formulation and closed-loop control}
\label{sec:formulation}

Let $V_{\mathrm{pre}}$ denote a predeformation OCT scan and $G=\mathcal T(V_{\mathrm{pre}};\eta)$ as a target volume constructed with deformation parameters $\eta$. The target remains fixed within a rollout. 
At step $t$, the policy receives the target, an observation history $V_{t-H_o+1:t}$, and the corresponding tool poses $\mathbf p_{t-H_o+1:t}$ to predict
\begin{equation}
 \widehat{\mathbf A}_t
 =\pi_\theta\!\left(V_{t-H_o+1:t},G,\mathbf p_{t-H_o+1:t}\right)
 \in\mathbb R^{H_a\times6}.
 \label{eq:policy-contract}
\end{equation}
Here $H_o$ is the observation horizon and $H_a$ is the action prediction horizon. 
Each row of $\widehat{\mathbf A}_t$ contains normalized translation and rotation increments of the tooltip pose (policy in Fig.~\ref{fig:closeloopsys}). 
These intermediate poses are defined relative to the initial tooltip pose, while action increments are executed in the current tooltip frame.
The predicted action sequence is clipped component-wise to $[-1,1]$ in normalized action space. 
The first action is then denormalized to translational increments in mm for $(\Delta x,\Delta y,\Delta z)$ and rotational increments in degrees for $(\Delta\phi,\Delta\theta,\Delta\psi)$, and executed.
A new OCT scan then updates the observation history for the next prediction. This receding-horizon procedure closes the loop between observed deformation and tooltip motion~\citep{chi2023diffusionpolicy}. 

\subsection{OCT representations}
\label{sec:representations}

We represent current OCT observations and target $G$ as full intensity volumes, tissue-masked sparse volumes, or voxelized point clouds. Each policy uses the same representation, spatial calibration, and preprocessing for current and target states.
These representations retain different combinations of OCT appearance and tissue geometry, enabling us to study how representation affects target-conditioned policy learning.

\noindent
\textbf{Full Volume:}
This representation retains the dense OCT intensity volume as an \(H \times W \times D\) voxel grid, preserving tissue structure and surrounding background signals.
A 3D convolutional encoder processes the complete volume to learn spatial and intensity features from the OCT observation.

\noindent
\textbf{Tissue-masked Sparse Volume:}
This representation concatenates the coordinates and intensity of each masked tissue voxel to form its input features.
A sparse convolutional encoder aggregates these features over neighborhoods defined by integer grid coordinates and uses global mean pooling to obtain a tissue-masked embedding (Appendix~\ref{app:architecture}).

\noindent
\textbf{Tissue-masked Voxelized Point Cloud:}
This representation retains the tissue voxel coordinates within the 3D OCT image space as an \(M \times 3\) point set, \((x,y,z)\), including the tissue interior while discarding OCT intensity information.
A PointNet-style encoder~\citep{qi2017pointnet} applies shared pointwise transformations followed by symmetric pooling to obtain a geometry-only embedding.

\subsection{Representation-specific residual conditioning}
\label{sec:conditioning}

Beyond the choice of OCT representation, we investigate how current and target tissue states are conditioned for action prediction.
All policies receive features from both the current observation and target.
Residual-conditioned variants additionally encode an explicit target--current discrepancy for each history observation, providing the policy with a direct representation of the remaining deformation.
For volumetric representations, this residual is computed on the spatial voxel lattice, whereas for voxelized point clouds, it is computed in the learned feature space.

\noindent
\textbf{Full-volume Residual:}
For normalized current and target volumes $\widetilde V_t$ and $\widetilde G$, the residual is the voxelwise difference $R_t=\widetilde G-\widetilde V_t$, which is encoded as an additional volume.

\noindent
\textbf{Sparse-tissue Residual:}
Let $M_t$ and $M_g$ denote binary occupancy masks on the same 3D voxel grid, and let $I_t$ and $I_g$ denote intensities, set to zero outside their respective masks. 
On the coordinate union $\mathcal U_t=\{\mathbf q:M_t(\mathbf q)=1\ \text{or}\ M_g(\mathbf q)=1\}$, the residual has two channels,
\begin{equation}
 \mathbf r_t(\mathbf q)=
 \begin{bmatrix}
 M_g(\mathbf q)I_g(\mathbf q)-M_t(\mathbf q)I_t(\mathbf q)\\
 M_g(\mathbf q)-M_t(\mathbf q)
 \end{bmatrix},\qquad \mathbf q\in\mathcal U_t.
 \label{eq:sparse-discrepancy}
\end{equation}
The intensity channel captures local OCT appearance differences within the masked volumes, while the occupancy channel explicitly encodes geometric mismatch where tissue is present in only one state.
Together, they distinguish changes in tissue appearance from changes in tissue occupancy, and a sparse encoder extracts features from this two-channel residual.

\noindent
\textbf{Voxelized Point Cloud Residual:}
For voxelized point clouds $P_t$ and $P_g$, a shared encoder $E$ produces pooled features $\mathbf z_t=E(P_t)$ and $\mathbf z_g=E(P_g)$.
We define the residual as $\mathbf r_t=\mathbf z_g-\mathbf z_t$, providing target--current discrepancy conditioning in feature space without requiring pointwise correspondence~\citep{thach2022deformernet}.

\begin{figure}[!htbp]
 \centering
 \includegraphics[width=\linewidth]{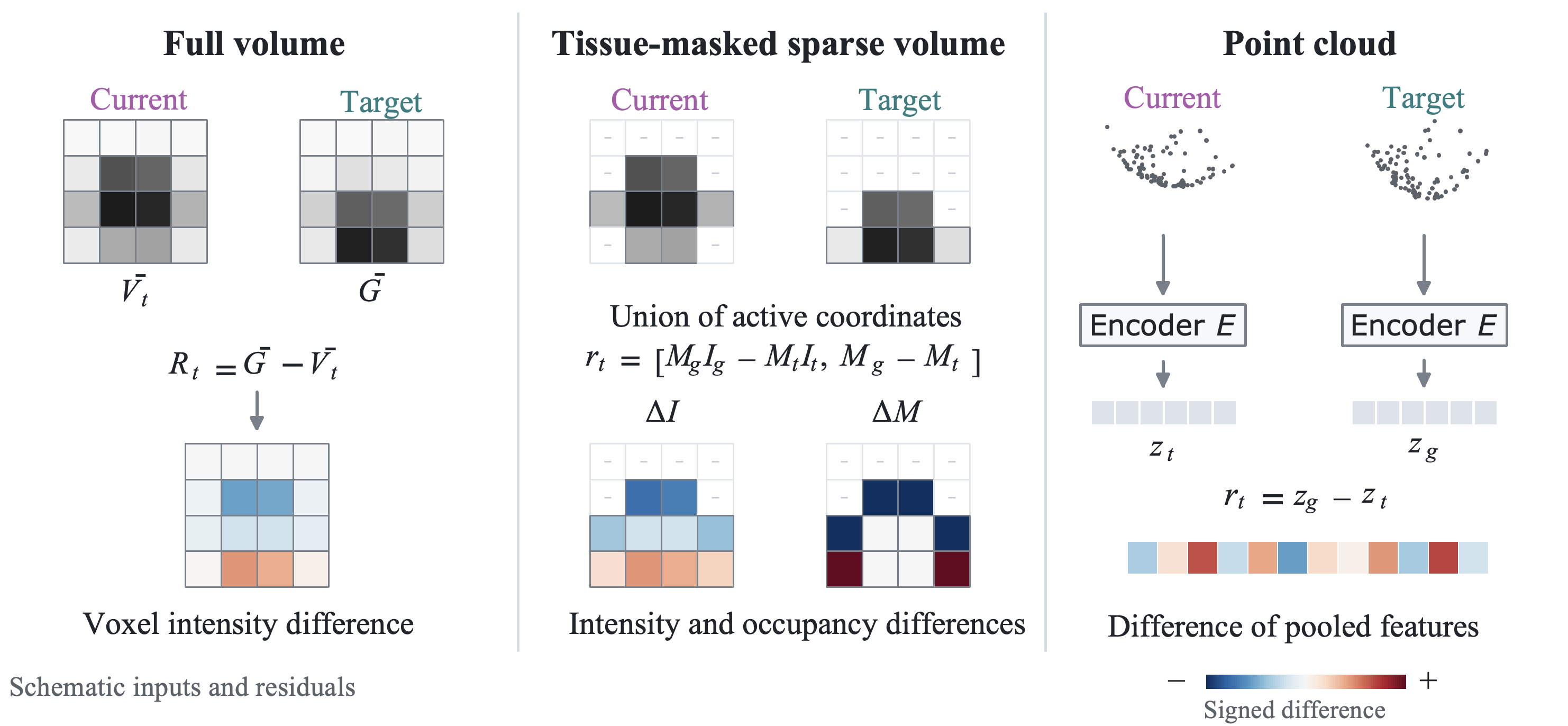}
 \caption{\textbf{Representation-specific residual conditioning.} 
  Full volumes use voxelwise intensity differences, sparse tissue volumes use intensity and occupancy differences over occupied coordinates, and voxelized point clouds use pooled feature differences. Values are schematic.
 }
 \label{fig:residual_conditioning}
\end{figure}

\subsection{Action sequence prediction}
\label{sec:policy-learning}

Observation and target features, relative tool poses, and optional residual features are combined into the conditioning input $c_t$.
Conditioned on $c_t$, each policy predicts a sequence of $H_a$ normalized 6-DoF tool-motion increments.
We compare deterministic behavioral cloning (Explicit) with diffusion-based policies that model conditional distributions over action sequences.

\noindent
\textbf{Explicit Policy:}
We adopt a deterministic explicit policy~\citep{chi2023diffusionpolicy} that maps the conditioning features $c_t$ directly to an action sequence.
A set of $H_a$ learned action queries attends to the conditioning features through a multi-head cross-attention decoder.
The decoder jointly predicts the full action sequence in a single forward pass, $\widehat{\mathbf A}_t=f_\theta^{\mathrm{explicit}}(c_t)$.

\noindent 
\textbf{Diffusion Policy:}
We train a conditional diffusion model~\citep{chi2023diffusionpolicy} to recover expert actions from noisy sequences.
During training, a normalized expert action sequence $\mathbf A_t$ is corrupted with Gaussian noise,
\begin{equation}
 \mathbf A_t^{(k)}
 =\sqrt{\bar\alpha_k}\,\mathbf A_t
 +\sqrt{1-\bar\alpha_k}\,\boldsymbol\epsilon,
 \qquad \boldsymbol\epsilon\sim\mathcal N(0,I),
 \label{eq:action-corruption}
\end{equation}
where $k$ denotes the diffusion step and $\bar{\alpha}_k$ controls the mixture of the expert action sequence and Gaussian noise.
Conditioned on the noisy action sequence, $k$, and $c_t$, a Transformer predicts the clean action sequence using self-attention over action tokens and cross-attention to the conditioning features.
At inference, the policy starts from Gaussian noise and iteratively denoises the action sequence through successive clean-action predictions.

\noindent
\textbf{Training and model selection.}
Both policy types are trained to reconstruct normalized expert action sequences. 
Let $\mathcal J_t$ contain the available future-action positions in a demonstration sequence. 
The training loss is
\begin{equation}
 \ell_t
 =\frac{1}{6|\mathcal J_t|}
 \sum_{j\in\mathcal J_t}
 \left\|\widehat{\mathbf A}_t[j]-\mathbf A_t[j]\right\|_2^2.
 \label{eq:masked-action-loss}
\end{equation}
The loss averages squared prediction error over valid future steps and all six action components. 
For diffusion policies, it is additionally averaged over sampled diffusion steps and noise realizations. 
Unavailable future actions are zero-padded and do not directly contribute to the loss.

All configurations share the demonstration split and closed-loop execution protocol in Section~\ref{sec:formulation}.
For each model, we select the checkpoint with the lowest validation action MSE in normalized space, with equal weighting across trajectories (implementation and training details in Appendix~\ref{app:implementation}).
\section{Real-Robot Evaluation}
\label{sec:experiments}

We evaluated the proposed framework on physical tissue phantoms and \textit{ex vivo} porcine tissue to determine whether learned policies (i) achieve desired tissue morphologies, (ii) adapt actions to different targets, and (iii) generalize to intermediate targets unseen during training.

\subsection{Experimental setup}
\label{sec:protocol}

The platform combines a 6-DoF UR5e robot, a printed palpation tooltip, a 1064-nm swept-source OCT system operating at 400 kHz, and an RGB camera (Fig.~\ref{fig:closeloopsys}).
Experiments use a soft silicone 40A phantom with an internal empty layer, together with \textit{ex vivo} porcine tissue.
Two operators collected 40 depth and 51 directional phantom demonstrations and 30 demonstrations per porcine task, totaling 151 trajectories and 1,714 OCT observations.
All configurations use an approximately 80/20 trajectory-level training/validation split (data acquisition and preprocessing details in Appendix~\ref{app:implementation}).

The phantom depth task evaluates 2--5 mm targets in 0.5-mm increments, with demonstrations at integer depths. The directional task evaluates $0^\circ$--$180^\circ$ in $22.5^\circ$ increments, with demonstrations every $45^\circ$, a depth of 3.5\,mm, and tilt strength of 0.7. Both use a target width of 6 mm.
Twelve policy configurations combine three OCT representations (full volume, tissue-masked sparse volume, and voxelized point cloud), with or without residual conditioning, and Explicit or Diffusion prediction.
Each checkpoint is evaluated over three target sweeps, yielding 252 depth and 324 directional trials. Between sweeps, the phantom is repositioned and rescanned, with deployment seeds varied where applicable.
Each rollout executes 12 actions with OCT feedback toward a fixed target generated from the initial scan. Repeated sweeps quantify deployment variability for each checkpoint.

\subsection{Evaluation metrics}
\label{sec:evaluation_metrics}

\noindent
\textbf{Shape accuracy:}
Let $P_t$ and $Q$ denote occupied voxel locations in the observed and target tissue masks, in calibrated millimeters. With $d(\mathbf p,Q)=\min_{\mathbf q\in Q}\|\mathbf p-\mathbf q\|_2$, symmetric Shape RMSE is
\begin{equation}
 D_t=\sqrt{\frac{1}{2|P_t|}\sum_{\mathbf p\in P_t}d(\mathbf p,Q)^2+
 \frac{1}{2|Q|}\sum_{\mathbf q\in Q}d(\mathbf q,P_t)^2}.
 \label{eq:geometric-rmse}
\end{equation}
The observation-to-target component ($P_t\!\to Q$) measures how far observed tissue lies outside the target support, whereas the target-to-observation component ($Q\!\to P_t$) measures how far uncovered target regions lie from the observed tissue.
Evaluation uses the native OCT grid without registration or policy-input resampling. 
We report final error $D_{12}$, reduction $\rho=100(1-D_{12}/D_0)$, and the number of rollouts satisfying $D_{12}<D_0$. 
Means and SDs summarize three target-averaged sweeps (HD95 and seen/unseen results in Appendix~\ref{app:phantom_protocol}).

\noindent
\textbf{Target-dependent action response and demonstration agreement:}
Distance correlation $r_d$ measures association between target parameters and composed endpoint commands; response spread $W$ measures their across-target variation. Larger $r_d$ indicates stronger association; $W$ describes the magnitude of across-target variation.
Separately, nearest-demonstration distance compares component-wise sums of command increments with the closest of ten demonstrations at the same target, independently for translation and rotation. It uses demonstrated targets only: four depths and five directions. Lower distance indicates closer command agreement (details in Appendix~\ref{app:demonstration_comparison}).

\subsection{Cross-task shape accuracy and target interpolation}
\label{sec:shape_results}

\begin{figure}[htbp]
 \centering
 \includegraphics[width=\linewidth]{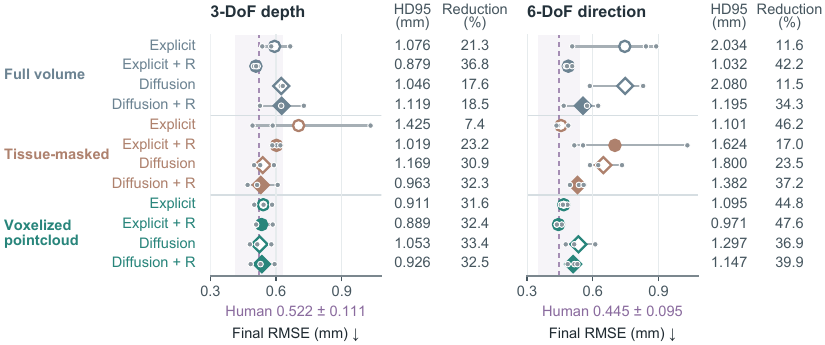}
 \caption{\textbf{Phantom shape accuracy and reliability.} 
 Large symbols show mean final shape RMSE; three gray dots and their range show target-averaged deployment sweeps. 
 Adjacent columns report mean final HD95 (details in Appendix~\ref{app:results}) and per-rollout RMSE reduction. 
 Purple dashed lines and bands show human endpoint means and reported $\pm1$ sample SD of final Shape RMSE across individual human demonstrations. Circles: Explicit; diamonds: Diffusion; filled: residual conditioning ($R$).}
 \label{fig:phantom_accuracy_summary}
\end{figure}

We first evaluate closed-loop shape accuracy, target generalization, and deployment repeatability on phantoms.
Closed-loop control reduces Shape RMSE in 512/576 phantom trials: 219/252 depth and 293/324 directional rollouts (Fig.~\ref{fig:phantom_accuracy_summary}). 
All four voxelized point-cloud configurations improve in at least 19/21 depth trials and all 27 directional trials, totaling 187/192, indicating improvement across action models and conditioning choices.
The lowest-error configuration differs by task: full-volume Explicit with residuals achieves the lowest depth error ($0.508\pm0.008$ mm), while voxelized point-cloud Explicit with residuals achieves the lowest directional error ($0.443\pm0.013$ mm) and a depth error of $0.534\pm0.044$ mm. The latter provides strong cross-task performance, with mean paired error reductions of 32.4\% and 47.6\% for depth and direction, respectively.
Voxelized point-cloud Explicit with residuals maintains accuracy at intermediate targets, with unseen versus seen errors of 0.537 versus 0.531 mm for depth and 0.427 versus 0.457 mm for direction, supporting interpolation within the tested target families.
Across three deployment sweeps, its target-averaged final RMSE ranges 0.508--0.584 mm for depth and 0.435--0.458 mm for direction, supporting repeatable closed-loop control across phantom repositioning and newly acquired predeformation scans.

\subsection{Effect of residual conditioning}
\label{sec:residual_results}

Residual conditioning lowers final RMSE in 9/12 matched comparisons, with the largest gains in full-volume directional control and recovery of target-dependent responses. 
Full-volume Explicit improves from 0.746 to 0.488 mm and Diffusion from 0.748 to 0.555 mm, with improving-rollout counts increasing from 14/27 to 27/27 and 19/27 to 26/27, respectively. 
Voxelized point-cloud Explicit shows smaller gains, from 0.541 to 0.534 mm for depth and 0.467 to 0.443 mm for direction.

\begin{figure}[htbp]
 \centering
 \includegraphics[width=\linewidth]{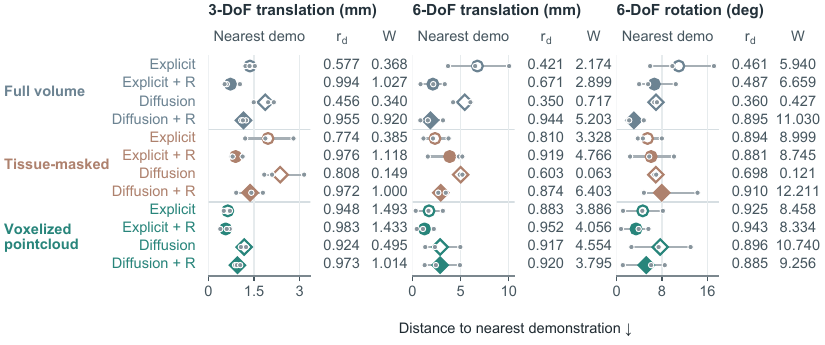}
 \caption{
 \textbf{Target response and agreement with human demonstrations.}
 Nearest-demonstration distances (``Nearest Demo'') show the mean across three deployment sweeps (various colors), with sweep means and ranges in gray. 
 Columns report pooled $r_d$ and $W$ across all targets, while demonstration distances use seen targets only. 
 Units apply to distances and $W$; symbols follow Fig.~\ref{fig:phantom_accuracy_summary}.
 }
 \label{fig:phantom_response_summary}
\end{figure}

Figure~\ref{fig:phantom_response_summary} separates target responsiveness from agreement with demonstrated commands. For Tissue-masked Diffusion in directional control, adding residuals increases translation spread from 0.063 to 6.403\,mm, which means residual conditioning produces more distinct target-dependent responses. This change accompanies stronger target association (0.603 to 0.874) and lower Shape RMSE (0.648 to 0.530\,mm). In depth control, nearest-demonstration distance decreases in all six matched pairs.

Stronger target responses do not necessarily improve shape accuracy. Full-volume Diffusion's depth association increases from 0.456 to 0.955 and spread from 0.340 to 0.920 mm, while RMSE remains nearly unchanged (0.623 versus 0.625 mm). For tissue-masked Explicit, directional translation association rises from 0.810 to 0.919, but nearest-demonstration distance increases from 2.337 to 3.859 mm and shape error from 0.455 to 0.701 mm. Thus, residuals can strengthen target-dependent commands, but their geometric benefit depends on the resulting tissue deformation.

\subsection{Explicit policy versus diffusion-based action prediction}
\label{sec:action_model_results}

Diffusion has lower depth RMSE in three of six matched comparisons, while Explicit has lower directional RMSE in five of six. With point-cloud residual inputs, the depth means of the two action models are nearly equal (0.534 versus 0.535\,mm), but Explicit has lower directional error (0.443 versus 0.510\,mm) and smaller nearest-demonstration distances in all three command measures. 
For full-volume residual inputs, Diffusion instead has closer directional command agreement, while Explicit achieves lower shape error (0.488 versus 0.555\,mm). 
Action-model choice therefore interacts with the representation and evaluation criterion. 
For voxelized point-cloud inputs with residuals, Explicit policy combines comparable depth accuracy, lower directional error, and faster prepared-input inference: 2.42\,ms versus 94.26\,ms for Diffusion on the timed depth checkpoints (Appendix~\ref{app:runtime}).

\subsection{Ex vivo porcine demonstration study}
\label{sec:exvivo}

To extend the phantom evaluation to biological tissue, we evaluate the framework on \textit{ex vivo} porcine tissue.
Separately trained full-volume Diffusion and voxelized point-cloud Explicit policies, both with residual conditioning, are evaluated at five depths (2--4\,mm) and five directions ($0^\circ$--$90^\circ$) over three deployment sweeps, totaling 60 rollouts.
Target width is 3\,mm; directional targets use depth 5\,mm and tilt strength 0.7. Intermediate depths (2.5, 3.5\,mm) and directions ($22.5^\circ$, $67.5^\circ$) are absent from demonstrations.

\begin{figure}[h]
 \centering
 \includegraphics[width=\linewidth]{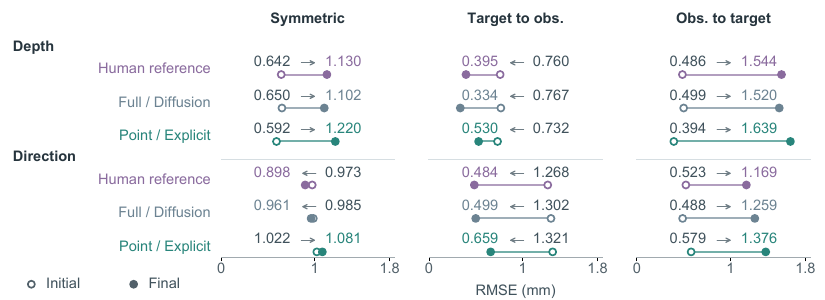}
 \caption{
 \textbf{Porcine tissue shape error before and after control.}
 Open and filled markers show initial and final mean RMSE, respectively; arrows indicate the direction of error change after control.
 Symmetric denotes Shape RMSE combining both distance directions; obs.\ denotes observed tissue geometry.
 Human references include 30 demonstrations per task, showing policy and human errors on a similar scale and with similar trends across conditions, as indicated by arrow orientation.
 }
 \label{fig:exvivo_summary}
\end{figure}

Figure~\ref{fig:exvivo_summary} compares policy outcomes with human demonstration references.
For depth control, final Shape RMSEs are 1.102 and 1.220 mm for the two policies versus 1.130 mm for human demonstrations.
For directional shaping, the corresponding errors are 0.961 and 1.081 mm versus 0.898 mm for human demonstrations.
Despite the complex \textit{ex vivo} tissue geometry, policy and human demonstration errors show similar trends across most conditions and remain on a similar scale.
Both policies also show strong target-dependent directional command responses (Appendix Table~\ref{tab:exvivo_target_response}).

\begin{figure}[htbp]
 \centering
 \includegraphics[width=0.95\linewidth]{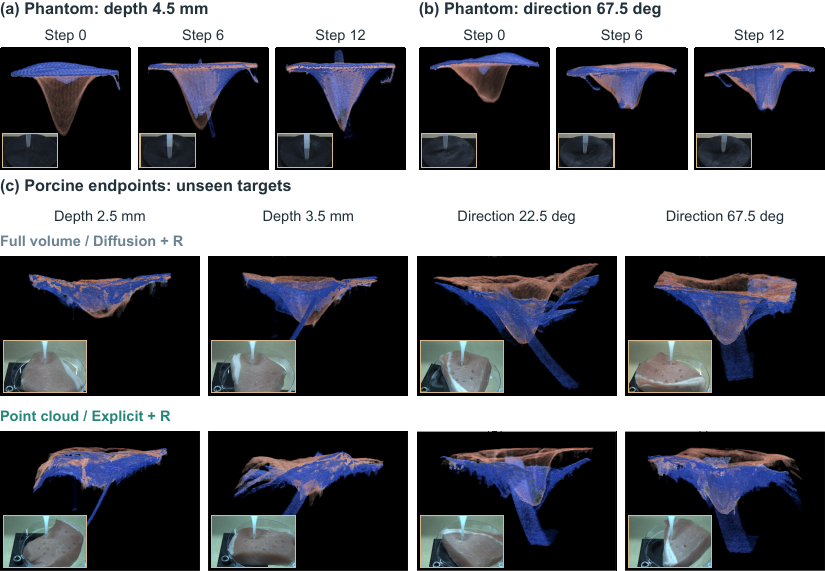}
 \caption{\textbf{Phantom deformation states and porcine endpoint geometry.} (a,b) OCT-derived geometry at initial, intermediate, and final phantom states. (c) Porcine endpoints at unseen targets. Blue: observed tissue; brown: fixed target; insets: RGB observations. The porcine overlays show remaining differences in depression width and surrounding geometry.}
 \label{fig:closed_loop_progression}
\end{figure}

Target-to-observation RMSE decreases in 58/60 policy rollouts and all 60 human demonstrations, whereas observation-to-target RMSE increases in every trajectory in both cohorts (Table~\ref{tab:exvivo-directed}). 
This paired pattern reflects coupled tissue deformation: approaching the target depression can simultaneously alter its width, sidewalls, and surrounding geometry, as illustrated in Fig.~\ref{fig:closed_loop_progression}c.

\section{Discussion and Conclusion}
\label{sec:conclusion}

In this work, we develop an OCT-guided, data-driven framework for closed-loop tissue shape control, which uses volumetric OCT to represent current and target tissue states and learns incremental tool motions from user-guided demonstrations.
Across full-volume, tissue-masked sparse-volume, and voxelized point-cloud representations, OCT supports both deterministic (Explicit) and generative diffusion-based (Diffusion) action prediction.
Voxelized point-cloud Explicit with residual conditioning provides accurate shape matching, performance at intermediate unseen targets, and repeatable deployment, while residual conditioning substantially improves full-volume policies with both action models.
Porcine experiments further extend the framework to biological tissue, where learned policies produce target-dependent deformation with shape errors on a similar scale to human demonstrations.

However, several limitations remain.
Model inference requires approximately 182--403 ms for full-volume, 35--615 ms for tissue-masked sparse-volume, and 2--94 ms for voxelized point-cloud representations across Explicit and Diffusion configurations (Appendix Table~\ref{tab:model-runtime}). 
Combined with OCT acquisition and robot execution latency, this cost limits the closed-loop update rate.
The current experiments also focus on short-horizon, localized tissue shaping and do not address longer interventions requiring temporal context and coordinated subtasks. 
Future work will investigate more efficient OCT representations and inference, long-horizon control, and anatomical and task-specific constraints for target generation. 
In summary, these results provide a foundation for integrating volumetric medical imaging, goal-conditioned learning, and generative action policies for more complex image-guided tissue manipulation.

\subsection*{AI use statement}

AI tools were used to assist with language editing and identifying relevant literature. 
The scientific content and initial manuscript drafts were developed by the authors, and all suggested references were independently reviewed and verified for relevance and accuracy. 
The authors reviewed all AI-assisted content and take full responsibility for the final text, claims, citations, and results.

\subsection*{Ethics statement}

This work evaluates robotic tissue manipulation using physical tissue phantoms and \textit{ex vivo} porcine tissue. 
The study focuses on engineering development of OCT-guided robotic manipulation with data collected from user demonstrations.
All experimental procedures and data collection were conducted in accordance with applicable institutional research and safety requirements.

\subsection*{Reproducibility statement}

The Appendix provides the evaluation protocol, target conditions, deployment procedure, and metrics, together with additional results and analyses.

\bibliography{references}
\bibliographystyle{oct_preprint}

\appendix
\section{Implementation and Reproducibility}
\label{app:implementation}

\subsection{Demonstrations and dataset partitions}
\label{app:data}
\label{app:dataset}

In this study, we constructed the training and validation datasets from user-guided manipulation demonstrations.
Each demonstration is one complete target-conditioned manipulation trajectory,
\begin{equation}
\mathcal D_i=
\left(
G_i,\,
\{V_{i,t}\}_{t=0}^{L_i},\,
\{\mathbf p_{i,t}\}_{t=0}^{L_i},\,
\{\mathbf a_{i,t}\}_{t=0}^{L_i-1}
\right),
\label{eq:demonstration-elements}
\end{equation}
where $G_i$ is the fixed target volume, $V_{i,0}$ is the predeformation OCT scan, and $V_{i,1:L_i}$ are the post-action OCT observations. The six-dimensional vector $\mathbf p_{i,t}$ describes the tooltip pose relative to its initial pose, using translation in millimetres and a rotation vector in degrees. The action $\mathbf a_{i,t}\in\mathbb R^6$ is the local tooltip command applied at state $t$, comprising three translational increments in millimetres and three extrinsic Euler-$xyz$ rotational increments in degrees. And each trajectory with $L_i$ actions contains $L_i+1$ observations and poses. The recorded trajectories may contain 9--15 actions.

Table~\ref{tab:demonstration-inventory} summarizes 151 demonstrations, 1,563 actions, and 1,714 OCT observations, including the initial scans and excluding the target volumes. Complete trajectories are split approximately 80/20 within each target condition using seed 42. Evaluation uses separately collected closed-loop rollouts, including intermediate target conditions absent from both training and validation.

\begin{table}[htbp]
\centering\small
\caption{Complete demonstration inventory and trajectory-level partitions.}
\label{tab:demonstration-inventory}
\begin{tabular}{lrrrr}
\toprule
Dataset & Demonstrations & Train/validation & Actions & Observations\\
\midrule
Phantom depth &40&32/8&405&445\\
Phantom directional &51&41/10&535&586\\
Porcine depth &30&24/6&302&332\\
Porcine directional &30&24/6&321&351\\
\midrule
Total &151&121/30&1,563&1,714\\
\bottomrule
\end{tabular}
\end{table}

\subsection{OCT Parameterized target generation}
\label{app:target-generation}
\label{app:targets}
\label{sec:goals}

We construct target OCT volumes from predeformation scans by extracting visible tissue using adaptive A-scan boundary detection and spatial continuity constraints~\citep{sun2016octsegmentation}. Connectivity filtering and smoothing refine the tissue surface, which is then deformed by a parameterized displacement field. Measured OCT intensities are resampled into the target geometry, preserving specimen-specific structure while enabling controlled target shapes.

Specifically, let $\mathbf{x}$ denote lateral position in millimetres, $\mathbf{c}_b$ the center of the OCT field of view, $A$ the nominal peak deformation depth, and $w$ the target base width. A depth-only target is defined by the Gaussian displacement field:
\begin{equation}
d_{\mathrm{depth}}(\mathbf{x})
=
A\exp\!\left(
-\frac{\|\mathbf{x}-\mathbf{c}_b\|_2^2}{2\sigma_w^2}
\right),
\qquad
\sigma_w
=
\frac{w}{2\sqrt{-2\log q_0}},
\qquad
q_0=0.05.
\label{eq:appendix-depth-target}
\end{equation}
Here, $A$ is expressed in millimetres, and $w$ is the diameter of the contour at which the displacement equals $0.05A$. The Gaussian tail is not truncated outside this contour.

Directional targets use two additional parameters: the direction $\phi$ and the tilt strength $\tau\in[0,1]$. The unit direction vector and displaced peak center are
\begin{equation}
\mathbf e_\phi =
\begin{bmatrix}
\sin\phi\\
-\cos\phi
\end{bmatrix},
\qquad
\mathbf c_p
=
\mathbf c_b+0.30\,\tau w\,\mathbf e_\phi.
\label{eq:target-peak-offset}
\end{equation}
Thus, $\tau=0$ leaves the peak at the scan center and recovers the depth-only target, whereas increasing $\tau$ moves the peak toward direction $\phi$.

To preserve a centered and smooth target base, only the inner Gaussian contours are shifted. For a normalized displacement level $q\in[q_0,1]$, the contour center and radius are
\begin{equation}
\mathbf c(q)
=
\mathbf c_b+
\frac{q-q_0}{1-q_0}
\left(\mathbf c_p-\mathbf c_b\right),
\qquad
r(q)=\sigma_w\sqrt{-2\log q}.
\label{eq:target-contours}
\end{equation}
Consequently, the outer $5\%$ contour remains centered at $\mathbf c_b$, while progressively higher contours move toward $\mathbf c_p$. Each lateral position is assigned the largest displacement level whose contour contains that position. This produces a smooth single-peaked directional deformation rather than translating the entire Gaussian footprint.

\begin{table}[htbp]
\centering
\small
\caption{Target parameters used for user demonstration collection. 
The dash denotes a depth-only target. Direction angles follow the native OCT row--column convention.}
\label{tab:target-settings}
\begin{tabular}{lcccc}
\toprule
Dataset & $A$ (mm) & $w$ (mm) & $\phi$ & $\tau$\\
\midrule
Phantom depth & $2,3,4,5$ & 6 & -- & --\\
Phantom directional & 3.5 & 6 & $0,45,90,135,180^\circ$ & 0.7\\
Porcine depth & $2,3,4$ & 3 & -- & --\\
Porcine directional & 5.0 & 3 & $0,45,90^\circ$ & 0.7\\
\bottomrule
\end{tabular}
\end{table}

Tissue displacement exposes regions without corresponding source-tissue samples. These regions are filled with measured air samples from the predeformation scan rather than zeros or synthetic Gaussian noise, preserving the measured OCT background distribution and avoiding artificial empty bands around the displaced tissue. The generated target retains the native OCT grid and intensity type and remains fixed throughout the demonstration and closed-loop rollout.

\begin{figure}[!htbp]
 \centering
 \includegraphics[width=0.85\linewidth]{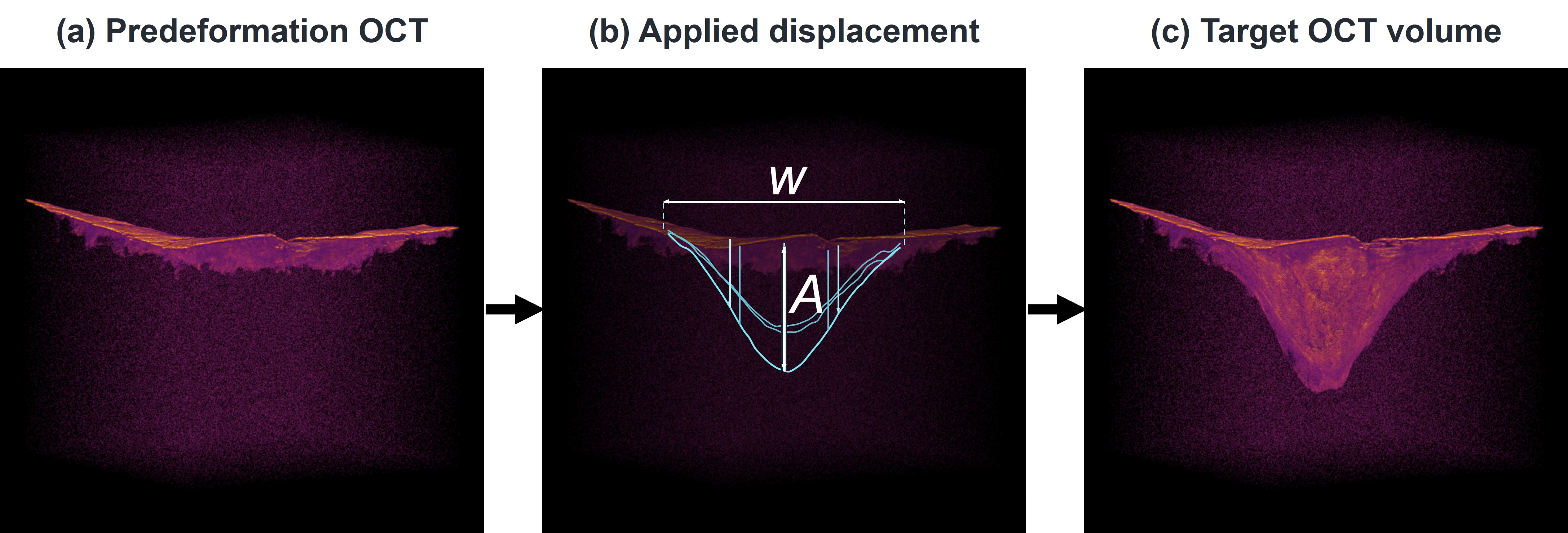}
 \caption{\textbf{Parameterized OCT target generation.} A measured predeformation OCT volume is transformed into a target tissue morphology using a parameterized displacement field with deformation depth $A$ and width $w$ defined in Equation~\ref{eq:appendix-depth-target}. The resulting target volume preserves the measured OCT structure within the deformed geometry.}
 \label{fig:oct_representations}
\end{figure}

\paragraph{Target Volume Construction:}
We construct the target by resampling the measured OCT intensities according to the deformation field (Fig.~\ref{fig:oct_representations}),
\begin{equation}
 G(\mathbf q)=V_{\mathrm{pre}}\!\left(\mathcal W_\eta(\mathbf q)\right),
 \label{eq:target-warp}
\end{equation}
where $\mathcal W_\eta$ maps target locations to sampling locations in the predeformation volume. 
For depth deformation, we define $\mathcal W_{A,w}(\mathbf x,z)=(\mathbf x,z-d_{\mathrm{depth}}(\mathbf x))$, so each target location samples the original volume at its axial position before displacement. Directional targets use the corresponding asymmetric displacement field. Resampling is applied within the deformed tissue region, and measured air samples fill newly exposed and out-of-bounds regions.

\subsection{OCT preprocessing and input representations}
\label{app:representations}
\label{app:preprocessing}

We used a \qty{1064}{\nano\meter}, \qty{400}{\kilo\hertz} swept-source OCT system (Thorlabs) with a fiber-based Mach--Zehnder interferometer for volumetric acquisition.
The interference signal was detected by a balanced photoreceiver and digitized at \qty{1}{\giga\sample\per\second} using an ATS-9364 digitizer (AlazarTech).

For the OCT data processing, the raw observations and target volumes are stored as signed \texttt{int8} arrays of shape $[1,256,256,768,1]$. After removing singleton dimensions, the array order is row--column--depth (HWD).
The lateral voxel spacing is 0.12857 mm for phantom scans and 0.05298 mm for \textit{ex vivo} scans, while the axial voxel spacing is 0.01325 mm and 0.01329 mm, respectively.

\paragraph{Artifact removal.}
A trajectory-specific artifact mask is estimated once from the predeformation scan and reused for every observation and the fixed target. We first form a coarse occupancy volume by thresholding local bright-voxel fractions and identify three-dimensional connected components. The scanner-top artifact is identified as the shallow component that touches the top boundary and spans most of the lateral field of view, and the resulting mask covers the leading axial band. For phantom data, the lower support plate is identified separately as a deeper component below the tissue that spans the lateral boundary and has a predominantly peripheral footprint. This classification is not applied to \textit{ex vivo} scans because the \textit{ex vivo} setup contains no lower support plate. Masked voxels are replaced with samples from measured predeformation air above the tissue surface, excluding both the scanner-top band and a guard interval immediately above the detected tissue surface.

\paragraph{Surface detection and visible-tissue support extraction.}
The same adaptive surface detector is used for both target generation and representation preprocessing. After artifact removal, the detector estimates the upper tissue surface in each A-scan from local background statistics by requiring a persistent axial intensity rise and retaining the largest laterally depth-consistent surface sheet. The reliable lateral footprint is then hole-filled and morphologically closed to suppress small gaps.

For target generation, the predeformation surface and its reliable footprint define the region over which the parameterized displacement field is applied. The measured OCT volume is inverse-warped according to this deformation, while newly exposed regions are filled with measured air sampled from the predeformation scan.

For the Tissue-masked Sparse Volume and Voxelized Point Cloud representations, we additionally construct a solid visible-tissue support mask. Within the reliable lateral footprint, OCT intensity is evaluated from the detected upper surface to approximately 300 axial indices below. At each lateral location, an adaptive threshold combines a fixed tissue-intensity criterion with measured-air statistics from the predeformation scan, and the deepest satisfying voxel defines the lower tissue boundary. All voxels between the upper and lower boundaries are labeled as tissue, forming a solid visible-tissue support region. This mask represents the OCT-supported visible tissue volume rather than only bright surface voxels or a complete anatomical segmentation. The procedure is applied independently to each current observation and target volume.

\paragraph{Normalization and coordinate convention:}
OCT intensities are normalized using the fixed global constants,
\begin{equation}
\widetilde I =
\operatorname{clip}\!\left(
\frac{I-\mu_I}{\sigma_I},-5,5
\right),
\qquad
\mu_I=10.377689,\quad
\sigma_I=6.639139.
\label{eq:oct-normalization}
\end{equation}
The same fixed transformation is used during training, validation, and
closed-loop inference. Volumes are converted from HWD to
depth--height--width (DHW) order and flipped along the established
scanner X axis. Observations and targets use the same cleanup,
normalization, orientation, and representation-specific resampling
procedure.

\paragraph{Full Volume:}
The Full Volume representation retains the cleaned tissue and measured air background as a dense OCT intensity volume. The normalized volume is trilinearly resampled to a fixed $256\times256\times256$ voxel grid. No tissue mask is provided as an additional input channel.

\paragraph{Tissue-masked Sparse Volume:}
The Tissue-masked Sparse Volume retains OCT intensities only within the visible-tissue mask. After resampling to a fixed $72\times128\times128$ DHW grid, each occupied voxel is represented by its integer grid coordinate, $\mathbf C\in\mathbb Z^{N\times3}$, and normalized OCT intensity, $\mathbf F\in\mathbb R^{N\times1}$, forming an $N\times4$ sparse representation. The number of occupied voxels $N$ varies with tissue occupancy. A sparse 3D convolutional encoder processes the occupied voxels and their local neighborhoods and pools them into a fixed 128-dimensional feature.

\paragraph{Voxelized Point Cloud:}
The Voxelized Point Cloud is derived from the same visible-tissue mask and retains geometry only. We sample 1,024 occupied voxel locations from the $72\times128\times128$ grid, reorder their DHW indices into XYZ coordinates, and normalize them to $[-1,1]^3$ using the fixed grid bounds. The resulting $1024\times3$ representation contains only normalized tissue voxel coordinates, without OCT intensity.

\begin{table}[htbp]
\centering
\small
\setlength{\tabcolsep}{4pt}
\caption{Policy inputs after preprocessing. $B$ denotes batch size; $N_t$ and $N_g$ denote the variable numbers of occupied sparse voxels in an observation and target, respectively. Sparse entries are shown per frame within one policy sample.}
\label{tab:input-contract}
\begin{tabular}{lll}
\toprule
Input & Observation history & Fixed target\\
\midrule
Full volume & $[B,3,256,256,256]$ & $[B,256,256,256]$\\
Tissue-only & Three $\{[N_t,3]_{\mathrm{coord}},[N_t,1]_{\mathrm{intensity}}\}$ sets & One $\{[N_g,3]_{\mathrm{coord}},[N_g,1]_{\mathrm{intensity}}\}$ set\\
Voxelized Point cloud & $[B,3,1024,3]$ & $[B,1024,3]$\\
Relative tool pose & $[B,3,6]$ & --\\
\bottomrule
\end{tabular}
\end{table}

\paragraph{Tooltip pose input and action units.}
At time $t$, the pose input is derived from the tooltip transform $T_{i,0}^{-1}T_{i,t}$ and represented by a three-dimensional translation in millimetres and a three-dimensional rotation vector in degrees. All six pose components are divided by 20 before entering the model. Action labels contain local translational increments in millimetres and extrinsic Euler-$xyz$ rotational increments in degree and all six action components are divided by 2.

\subsection{Representation encoders and action models}
\label{app:architecture}

All condition tokens have width 128. Table~\ref{tab:physical-architecture} summarizes the representation encoders and token counts. For Full Volume inputs, the three observations share one dense encoder, while the fixed target uses a separate encoder; residual conditioning adds a third encoder for residual volumes. For the Tissue-masked Sparse Volume, observations and target share a sparse encoder, with an additional sparse encoder for target--observation residuals. For the Voxelized Point Cloud, observations and target share a PointNet encoder, while residuals are computed by subtracting their embeddings.

A $6\to128\to128$ pose MLP with SiLU produces one token per history state. Temporal and token-type embeddings distinguish observation, target, pose, and residual tokens. Full Volume tokens additionally receive fixed 3D Fourier spatial features over the pooled $4^3$ token grid. Tissue-masked sparse-volume inputs use integer grid coordinates to locate occupied voxels and normalized OCT intensity as the voxel feature; voxelized point-cloud inputs use scanner-fixed normalized XYZ coordinates.

\begin{table}[htbp]
\centering
\small
\caption{Representation encoders and total condition-token counts without/with residual conditioning. Counts include three observations, one fixed target, and three relative tooltip poses.}
\label{tab:physical-architecture}
\begin{tabular}{p{0.17\linewidth}p{0.61\linewidth}r}
\toprule
Input & Encoder & Tokens\\
\midrule
Full volume &
Four dense 3-D convolutional residual stages with channels
$16/32/64/128$, GroupNorm, SiLU, and adaptive average pooling to
$4^3$ tokens per volume &
259 / 451\\

Tissue-only &
Sparse 3-D encoder with 27-neighbor blocks of channels
$32/64/128/128$ and strides $1/2/2/2$, followed by global mean
pooling over active voxels &
7 / 10\\

Voxelized point cloud &
Pointwise MLP $3\to64\to128\to128$ with ReLU, max pooling over
1,024 points, and pooled MLP $128\to128\to128$ &
7 / 10\\
\bottomrule
\end{tabular}
\end{table}

Each Full Volume input contributes $4^3=64$ tokens. Three observations, one target, and three pose tokens yield $3(64)+64+3=259$ tokens; three residual volumes increase the total to 451. Tissue-masked Sparse Volume and Voxelized Point Cloud inputs each produce one token per observation, target, or residual, yielding seven tokens without residual conditioning and ten with it.

\paragraph{Explicit Policy.}
The Explicit Policy is a deterministic direct action-regression decoder. Eight learned action queries, corresponding to the eight prediction-horizon steps, attend to the condition tokens through one four-head cross-attention layer. The attended features are added to the learned queries and passed through LayerNorm and an MLP ($128\to512\to6$, with SiLU), producing a normalized $8\times6$ action chunk in one forward pass.

\paragraph{Diffusion Policy.}
The Diffusion policy predicts the clean normalized $8\times6$ action chunk from a noisy sequence using a 100-step cosine diffusion schedule~\citep{nichol2021improved}. Noisy actions are projected to width 128 and combined with learned action-position embeddings. A diffusion-timestep embedding conditions four Transformer blocks, each with action self-attention, cross-attention to the condition tokens, four attention heads, and feed-forward width 512. Inference starts from Gaussian action noise and uses 20 deterministic DDIM-style updates~\citep{song2021ddim}.

Both policies use the same representation-specific condition tokens. 
Their normalized outputs are bounded to $[-1,1]$ and converted back to physical action units defined in Appendix~\ref{app:preprocessing}.
Fig.~\ref{fig:physical_policy_architecture} illustrates the shared conditioning pipeline, two action decoders, and closed-loop execution procedures.

\begin{figure}[htbp]
\centering
\includegraphics[width=\linewidth]{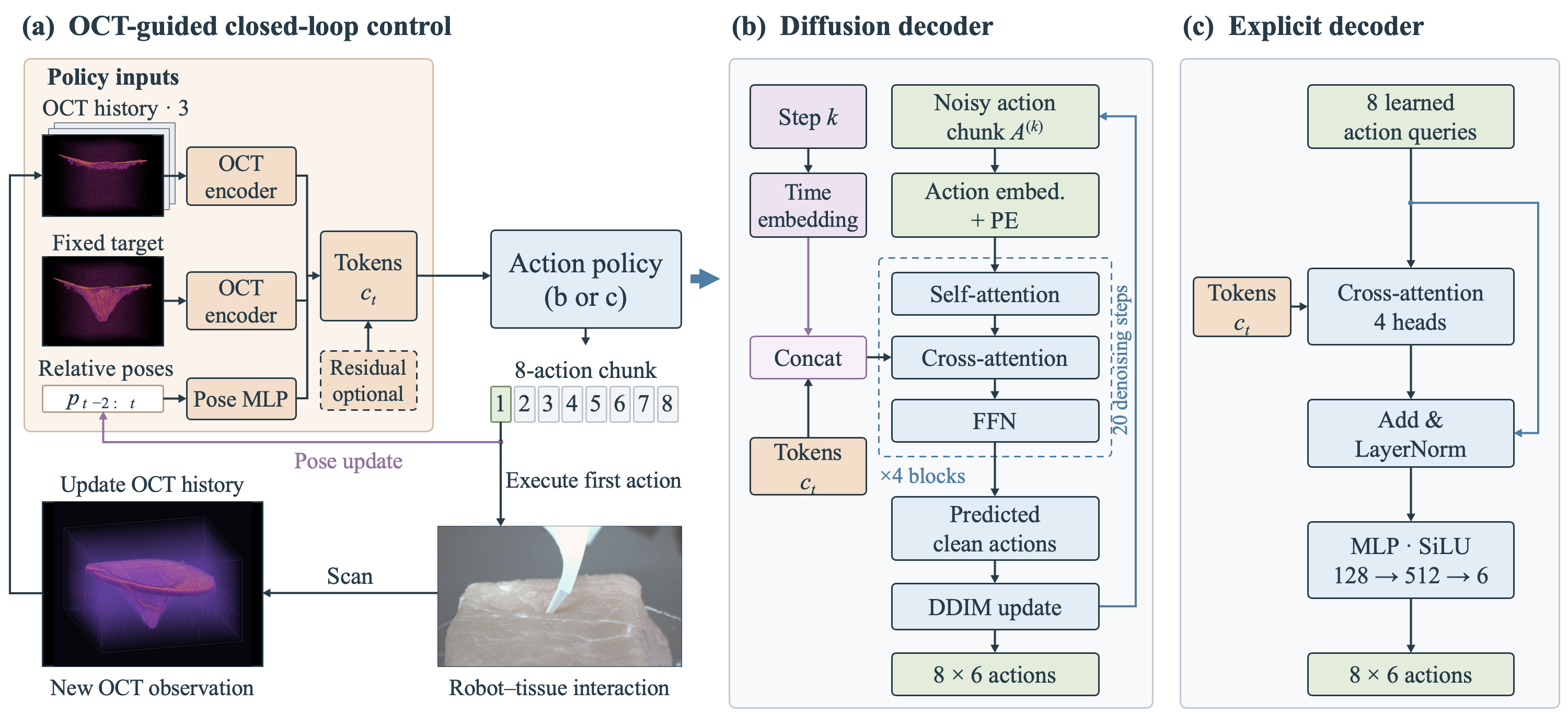}
\caption{\textbf{OCT-guided control and action decoders.} (a) OCT history, fixed targets, relative commanded poses, and optional residuals form condition tokens. The stack denotes three observations. Only the first of eight predicted actions is executed before rescanning; pose history is updated by command composition. (b) Four time-conditioned Transformer blocks predict clean actions for 20 DDIM-style updates. PE denotes learned action-position embeddings. (c) Explicit uses eight learned queries and one four-head cross-attention operation. Query skip addition, LayerNorm, and a SiLU MLP produce the action chunk. Representation encoders and token counts are specified in Table~\ref{tab:physical-architecture}.}
\label{fig:physical_policy_architecture}
\end{figure}

\subsection{Training and checkpoint selection}
\label{app:training}

\paragraph{Training samples and objective.}
We optimize all policies with AdamW~\citep{loshchilov2019decoupled} using the settings in Table~\ref{tab:training-settings}.
Each action defines a sample with history indices $[\max(0,t-2),\max(0,t-1),t]$ and up to eight future actions from $t$. The oldest observation and pose are repeated to initialize the history. Missing future actions are zero-padded and excluded from the masked six-dimensional MSE in Eq.~\eqref{eq:masked-action-loss}; no terminal-action samples are added.

\begin{table}[htbp]
\centering\small
\caption{Shared physical-policy training settings. The learning rate is constant.}
\label{tab:training-settings}
\begin{tabular}{ll@{\qquad}ll}
\toprule
Setting &Value&Setting&Value\\
\midrule
Optimizer &AdamW&Learning rate &$10^{-4}$\\
Weight decay &$10^{-4}$&Batch size &1\\
Maximum/minimum epochs &100/10&Early-stop patience &10\\
Minimum improvement &$10^{-4}$&EMA decay &0.995\\
Gradient accumulation &None&Training seed &42\\
\bottomrule
\end{tabular}
\end{table}

\paragraph{Checkpoint selection.}
Validation and deployment use EMA weights. For normalized expert commands $\bar{\mathbf a}_{i,t}=\mathbf a_{i,t}/2$, selection uses trajectory-macro first-action MSE:
\begin{equation}
 E_{\mathrm{val}}=\frac{1}{n_{\mathrm{val}}}\sum_{i=1}^{n_{\mathrm{val}}}
 \frac{1}{6L_i}\sum_{t=0}^{L_i-1}
 \|\widehat{\mathbf a}_{i,t,0}-\bar{\mathbf a}_{i,t}\|_2^2.
 \label{eq:physical-selection}
\end{equation}
Normalized MSE is averaged within each validation trajectory and then equally across trajectories. Diffusion validation generates one chunk per sample from pure noise, resetting the sampling seed to 42 for each validation pass. A checkpoint replaces the current best only when this score improves by more than $10^{-4}$. Early stopping uses the settings in Table~\ref{tab:training-settings}. We train and select one task-specific checkpoint for each configuration: twelve configurations for each phantom task and two for each porcine task, giving 28 evaluated checkpoints.

\subsection{Closed-loop execution and model-only runtime}
\label{app:execution-runtime}
\label{app:runtime}

The controller retains a fixed target and a three-state observation/pose history. It predicts eight actions, clips the normalized output to $[-1,1]$, multiplies the first action by two, and executes it. Component bounds are therefore $\pm2$\,mm and $\pm2^\circ$. The local tooltip increment is composed as $T_{t+1}=T_t\Delta T_t$ and converted to an end-effector command for inverse kinematics. A new OCT scan and the updated commanded pose enter the next prediction. Pose history is propagated from the initial robot pose by command composition. Each quantitative rollout executes 12 actions and stores every step scan.

\begin{table}[htbp]
\centering\small
\caption{Prepared-input model-only latency (mean $\pm$ sample SD, ms) for the twelve phantom depth checkpoints on an RTX 4500 Ada GPU: one fixed input per checkpoint, batch size one, FP16 autocast, five warm-ups, and 20 CUDA-synchronized calls.}
\label{tab:runtime}
\label{tab:model-runtime}
\begin{tabular}{lrr}
\toprule
Representation / conditioning &Explicit&Diffusion\\
\midrule
Full volume &$181.67\pm0.08$&$272.85\pm0.54$\\
Full volume + residual &$311.44\pm0.17$&$402.91\pm0.40$\\
Tissue-only &$35.06\pm1.44$&$140.16\pm14.73$\\
Tissue-only + residual &$517.86\pm77.71$&$614.60\pm39.28$\\
Point cloud &$2.16\pm0.23$&$91.16\pm0.57$\\
Point cloud + residual &$2.42\pm0.11$&$94.26\pm0.79$\\
\bottomrule
\end{tabular}
\end{table}

These values measure prepared-input model inference rather than the full control cycle. Timing includes encoders, residual operations, and the complete action head, including all 20 Diffusion updates and internal sparse bookkeeping. It excludes checkpoint loading, file I/O, preprocessing, representation extraction, OCT acquisition, inverse kinematics, and robot motion. 
\FloatBarrier

\section{Additional Real-Robot Results}
\label{app:results}
This appendix reports the complete numerical results of the real-robot evaluation, including shape accuracy, target-dependent action responses, agreement with demonstrations, and closed-loop progression. Geometric results use the native-grid evaluation of 576 phantom and 60 porcine rollouts. Command analyses retain the recorded phantom and porcine trajectories.

\subsection{Evaluation scope and reporting conventions}
\label{app:phantom_protocol}
\label{app:phantom}

\paragraph{Evaluation data and target conditions.}
A target value is \emph{seen} if it occurs in the corresponding demonstration training or validation data. \emph{Unseen} values are intermediate target parameters absent from both partitions. All evaluation rollouts are newly executed closed-loop deployments and are separate from the demonstrations used for training and validation.

The evaluation contains 636 twelve-action rollouts, 7,632 executed actions, and 8,268 OCT states, including the initial state of each rollout. Phantom evaluation contains 252 depth and 324 directional rollouts. Porcine evaluation contains 30 depth and 30 directional rollouts across the two tested configurations.

\begin{table}[t]
\centering
\small
\caption{Closed-loop evaluation inventory. Each phantom task evaluates all twelve policy configurations. Porcine experiments use task-specific checkpoints for the two configurations shown below. Each rollout contains twelve executed actions and thirteen OCT states, including the initial scan.}
\label{tab:results_evaluation_inventory}
\renewcommand{\arraystretch}{1.1}
\setlength{\tabcolsep}{4pt}
\begin{tabular*}{\linewidth}{@{\extracolsep{\fill}}lrrrr@{}}
\toprule
Task & Targets & Rollouts & Actions & OCT states \\
\midrule
\multicolumn{5}{l}{\textit{Phantom: all twelve configurations}} \\
Depth       & 7 & 252 & 3,024 & 3,276 \\
Directional & 9 & 324 & 3,888 & 4,212 \\
\midrule
\multicolumn{5}{l}{\textit{Porcine: Full-volume Diffusion $+$ residual}} \\
Depth       & 5 & 15 & 180 & 195 \\
Directional & 5 & 15 & 180 & 195 \\
\midrule
\multicolumn{5}{l}{\textit{Porcine: Point-cloud Explicit $+$ residual}} \\
Depth       & 5 & 15 & 180 & 195 \\
Directional & 5 & 15 & 180 & 195 \\
\midrule
\textbf{Total} & -- & \textbf{636} & \textbf{7,632} & \textbf{8,268} \\
\bottomrule
\end{tabular*}
\end{table}

\paragraph{Repeated deployments.}
Each phantom policy--target condition is evaluated in three independent deployment sweeps. Between sweeps, the phantom is repositioned to vary its placement height and orientation within the OCT field of view, a new predeformation scan is acquired, and the corresponding target is regenerated from that scan. Deployment seeds are also changed where applicable. 

For porcine evaluation, each configuration--target condition contains three independently executed rollouts. Because biological tissue cannot be reset to an identical initial state, these recordings are treated as repeated physical deployments rather than identical-state repeats.

\begin{table}[htbp]
\centering
\small
\caption{Phantom and porcine evaluation target conditions. Porcine experiments use task-specific checkpoints for Full-volume Diffusion $+$ residual and Point-cloud Explicit $+$ residual.}
\label{tab:evaluation_target_conditions}
\renewcommand{\arraystretch}{1.15}
\setlength{\tabcolsep}{4pt}
\begin{tabularx}{\linewidth}{
    @{} l l
    >{\raggedright\arraybackslash}X
    >{\raggedright\arraybackslash}p{0.25\linewidth}
    @{}
}
\toprule
Material & Task & Target conditions & Fixed parameters \\
\midrule

Phantom &
Depth &
\textbf{Seen:} $2.0,\,3.0,\,4.0,\,5.0$\,mm \newline
\textbf{Unseen:} $2.5,\,3.5,\,4.5$\,mm &
Width: $6$\,mm \\

Phantom &
Directional &
\textbf{Seen:} $0^\circ,\,45^\circ,\,90^\circ,\,135^\circ,\,180^\circ$ \newline
\textbf{Unseen:} $22.5^\circ,\,67.5^\circ,\,112.5^\circ,\,157.5^\circ$ &
Depth: $3.5$\,mm \newline
Width: $6$\,mm \newline
Strength: $0.7$ \\

\midrule

Porcine &
Depth &
\textbf{Seen:} $2.0,\,3.0,\,4.0$\,mm \newline
\textbf{Unseen:} $2.5,\,3.5$\,mm &
Width: $3$\,mm \\

Porcine &
Directional &
\textbf{Seen:} $0^\circ,\,45^\circ,\,90^\circ$ \newline
\textbf{Unseen:} $22.5^\circ,\,67.5^\circ$ &
Depth: $5$\,mm \newline
Width: $3$\,mm \newline
Strength: $0.7$ \\

\bottomrule
\end{tabularx}
\end{table}

\paragraph{How means and variability are computed.}
For each policy configuration $c$, let $m_{c,k,r}$ denote a rollout-level statistic at target condition $k$ in phantom deployment sweep $r$. For a reported all-target, seen-target, or unseen-target subset containing $K$ target values, we compute
\begin{equation}
\bar m_{c,r}=\frac{1}{K}\sum_{k=1}^{K}m_{c,k,r},
\qquad
\bar m_c=\frac{1}{3}\sum_{r=1}^{3}\bar m_{c,r},
\qquad
s_{m,c}=
\sqrt{\frac{1}{2}\sum_{r=1}^{3}
\left(\bar m_{c,r}-\bar m_c\right)^2}.
\label{eq:results-aggregation}
\end{equation}
Thus, statistics are reported as the mean $\pm$ sample SD of three target-averaged deployment sweeps for one fixed checkpoint.

\paragraph{Shape RMSE: geometric mismatch.}
We evaluate tissue--target alignment using the visible-tissue support masks defined in Section~\ref{app:preprocessing}. Let $P_t$ and $Q$ denote the sets of all occupied voxel locations within the observed and fixed-target support masks at step $t$. Each occupied voxel index is converted to physical coordinates using the material-specific lateral and axial voxel spacings reported in Section~\ref{app:preprocessing}. For a point $\mathbf p$ and point set $Q$, let
$d(\mathbf p,Q)=\min_{\mathbf q\in Q}\|\mathbf p-\mathbf q\|_2$.
We compute the two directed nearest-neighbor RMSEs and their symmetric combination:
\begin{equation}
\begin{gathered}
R_{P\rightarrow Q,t}
=
\sqrt{\frac{1}{|P_t|}
\sum_{\mathbf{p}\in P_t} d(\mathbf{p},Q)^2},
\qquad
R_{Q\rightarrow P,t}
=
\sqrt{\frac{1}{|Q|}
\sum_{\mathbf{q}\in Q} d(\mathbf{q},P_t)^2},\\
D_t
=
\sqrt{\frac{
R_{P\rightarrow Q,t}^{2}+R_{Q\rightarrow P,t}^{2}}{2}}.
\end{gathered}
\label{eq:results-shape-rmse}
\end{equation}
We report $D_t$ as Shape RMSE in millimeters; lower values indicate closer geometric agreement with the target. Evaluation is performed directly on the native OCT grid, without registration or policy-input resampling.

\paragraph{HD95: 95th-percentile geometric distance.}
Using all occupied voxel locations in the same visible-tissue support masks as Shape RMSE, we compute the 95th percentile of nearest-neighbor distances in each direction and report the larger value:
\begin{equation}
H_{95,t}=\max\!\left\{
\operatorname{quantile}_{0.95}\{d(\mathbf p,Q):\mathbf p\in P_t\},
\;
\operatorname{quantile}_{0.95}\{d(\mathbf q,P_t):\mathbf q\in Q\}
\right\}.
\label{eq:results-hd95}
\end{equation}
HD95 is reported in millimeters; lower values indicate closer point-set agreement at the 95th-percentile level. It complements Shape RMSE by describing the larger nearest-neighbor distances, while being less sensitive than the maximum distance to a small number of extreme values. 

\paragraph{Reduction and improved counts.}
For rollout $i$, we define the percentage reduction in Shape RMSE and an improvement indicator as
\begin{equation}
\rho_i=100\left(1-\frac{D_{i,12}}{D_{i,0}}\right),\qquad
I_i=\mathbf1[D_{i,12}<D_{i,0}].
\label{eq:results-reduction}
\end{equation}
$\rho_i$ measures the relative change from the initial to final Shape RMSE, and $\sum_i I_i$ reports the number of rollouts whose final Shape RMSE is lower than the initial value.

\paragraph{Endpoint results.}
Table~\ref{tab:results_phantom_endpoint} reports the final shape-matching results for all twelve configurations in the depth and directional tasks.

\begin{table}[h]
\centering
\footnotesize
\setlength{\tabcolsep}{2.4pt}
\renewcommand{\arraystretch}{1.10}
\caption{Complete phantom endpoint results evaluation. $D_{12}$ and $H_{95}$ are in millimeters; $\rho$ is the percentage reduction in Shape RMSE.}
\label{tab:results_phantom_endpoint}
\begin{tabular*}{\linewidth}{@{\extracolsep{\fill}}llrrrrrr@{}}
\toprule
Configuration & $R$ & $D_{12}$ all & $D_{12}$ seen & $D_{12}$ unseen & $H_{95}$ & $\rho$ & Improved \\
\midrule
\multicolumn{8}{l}{\textit{3-DoF depth}} \\
Full volume / Explicit & -- & $0.593\pm0.065$ & 0.602 & 0.581 & 1.076 & 21.3 & 17/21 \\
Full volume / Explicit & $\checkmark$ & $0.508\pm0.008$ & 0.499 & 0.520 & 0.879 & 36.8 & 21/21 \\
Full volume / Diffusion & -- & $0.623\pm0.006$ & 0.637 & 0.606 & 1.046 & 17.6 & 15/21 \\
Full volume / Diffusion & $\checkmark$ & $0.625\pm0.100$ & 0.636 & 0.612 & 1.119 & 18.5 & 17/21 \\
Tissue-only / Explicit & -- & $0.703\pm0.288$ & 0.710 & 0.693 & 1.425 & 7.4 & 16/21 \\
Tissue-only / Explicit & $\checkmark$ & $0.601\pm0.018$ & 0.590 & 0.616 & 1.019 & 23.2 & 18/21 \\
Tissue-only / Diffusion & -- & $0.539\pm0.046$ & 0.557 & 0.515 & 1.169 & 30.9 & 18/21 \\
Tissue-only / Diffusion & $\checkmark$ & $0.530\pm0.070$ & 0.533 & 0.526 & 0.963 & 32.3 & 18/21 \\
Point cloud / Explicit & -- & $0.541\pm0.041$ & 0.535 & 0.550 & 0.911 & 31.6 & 20/21 \\
Point cloud / Explicit & $\checkmark$ & $0.534\pm0.044$ & 0.531 & 0.537 & 0.889 & 32.4 & 20/21 \\
Point cloud / Diffusion & -- & $0.524\pm0.050$ & 0.536 & 0.509 & 1.053 & 33.4 & 20/21 \\
Point cloud / Diffusion & $\checkmark$ & $0.535\pm0.055$ & 0.543 & 0.525 & 0.926 & 32.5 & 19/21 \\
\midrule
\multicolumn{8}{l}{\textit{6-DoF directional}} \\
Full volume / Explicit & -- & $0.746\pm0.209$ & 0.750 & 0.742 & 2.034 & 11.6 & 14/27 \\
Full volume / Explicit & $\checkmark$ & $0.488\pm0.015$ & 0.502 & 0.472 & 1.032 & 42.2 & 27/27 \\
Full volume / Diffusion & -- & $0.748\pm0.140$ & 0.761 & 0.732 & 2.080 & 11.5 & 19/27 \\
Full volume / Diffusion & $\checkmark$ & $0.555\pm0.080$ & 0.570 & 0.537 & 1.195 & 34.3 & 26/27 \\
Tissue-only / Explicit & -- & $0.455\pm0.029$ & 0.466 & 0.440 & 1.101 & 46.2 & 27/27 \\
Tissue-only / Explicit & $\checkmark$ & $0.701\pm0.287$ & 0.707 & 0.694 & 1.624 & 17.0 & 18/27 \\
Tissue-only / Diffusion & -- & $0.648\pm0.076$ & 0.653 & 0.642 & 1.800 & 23.5 & 27/27 \\
Tissue-only / Diffusion & $\checkmark$ & $0.530\pm0.032$ & 0.538 & 0.521 & 1.382 & 37.2 & 27/27 \\
Point cloud / Explicit & -- & $0.467\pm0.015$ & 0.477 & 0.455 & 1.095 & 44.8 & 27/27 \\
Point cloud / Explicit & $\checkmark$ & $0.443\pm0.013$ & 0.457 & 0.427 & 0.971 & 47.6 & 27/27 \\
Point cloud / Diffusion & -- & $0.534\pm0.069$ & 0.546 & 0.519 & 1.297 & 36.9 & 27/27 \\
Point cloud / Diffusion & $\checkmark$ & $0.510\pm0.023$ & 0.513 & 0.507 & 1.147 & 39.9 & 27/27 \\
\bottomrule
\end{tabular*}
\end{table}

\subsection{Operator shape reference at demonstrated goals}
\label{app:human-baseline}
We evaluate the initial and final OCT states of all 151 operator demonstrations using the same material-specific visible-support extraction and geometric metrics as policy deployment. The 30 validation demonstrations serve as the primary operator reference; they are excluded from gradient updates and used only for checkpoint selection. The full 151-trajectory cohort summarizes all collected demonstrations. For matched target comparisons, the Point-cloud Explicit $+R$ policy includes only deployments at target parameters represented in the demonstrations. Each trajectory endpoint is its final recorded state, with goals weighted equally within each material and task. Operator and policy rollouts use distinct initial scans and may contain different numbers of actions.

\begin{table}[!htbp]
\centering\small
\setlength{\tabcolsep}{3pt}
\caption{Shape outcomes at demonstrated goal parameters. $D_0$ and $D_{\rm end}$ are initial and endpoint Shape RMSE; $H_{95,\rm end}$ is endpoint HD95 (all in mm). $n$ counts trajectories and ``Improved'' means $D_{\rm end}<D_0$. Operator trajectories end after 9--15 recorded actions; policy rollouts execute twelve. All and validation operator rows overlap; the latter are included in the former. Initial scans are distinct across cohorts.}
\label{tab:app-human-baseline}
\begin{tabular*}{\linewidth}{@{\extracolsep{\fill}}llrrrrr@{}}
\toprule
Material / task & Cohort & $n$ & $D_0\to D_{\rm end}$ & $H_{95,\rm end}$ & Improved & Actions\\
\midrule
Phantom / depth & Operators, all & 40 & $0.849\to0.522$ & 0.831 & 40/40 & 10--12 \\
Phantom / depth & Operators, val. & 8 & $0.849\to0.552$ & 0.835 & 8/8 & 10--12 \\
Phantom / depth & Point E+$R$ & 12 & $0.838\to0.531$ & 0.896 & 11/12 & 12 \\
\addlinespace[2pt]
Phantom / direction & Operators, all & 51 & $0.877\to0.445$ & 0.964 & 51/51 & 9--15 \\
Phantom / direction & Operators, val. & 10 & $0.876\to0.421$ & 0.953 & 10/10 & 10--13 \\
Phantom / direction & Point E+$R$ & 15 & $0.852\to0.457$ & 1.004 & 15/15 & 12 \\
\addlinespace[2pt]
Porcine / depth & Operators, all & 30 & $0.642\to1.130$ & 4.344 & 0/30 & 10--11 \\
Porcine / depth & Operators, val. & 6 & $0.623\to1.213$ & 4.769 & 0/6 & 10 \\
Porcine / depth & Point E+$R$ & 9 & $0.599\to1.256$ & 4.812 & 0/9 & 12 \\
\addlinespace[2pt]
Porcine / direction & Operators, all & 30 & $0.973\to0.898$ & 2.760 & 21/30 & 10--14 \\
Porcine / direction & Operators, val. & 6 & $0.998\to0.909$ & 2.600 & 6/6 & 10--14 \\
Porcine / direction & Point E+$R$ & 9 & $1.029\to1.088$ & 3.075 & 2/9 & 12 \\
\addlinespace[2pt]
\bottomrule
\end{tabular*}
\end{table}

\subsubsection{Shape RMSE over execution}
\label{app:rollout_curves}

\begin{figure}[h]
\centering
\includegraphics[width=0.95\linewidth]{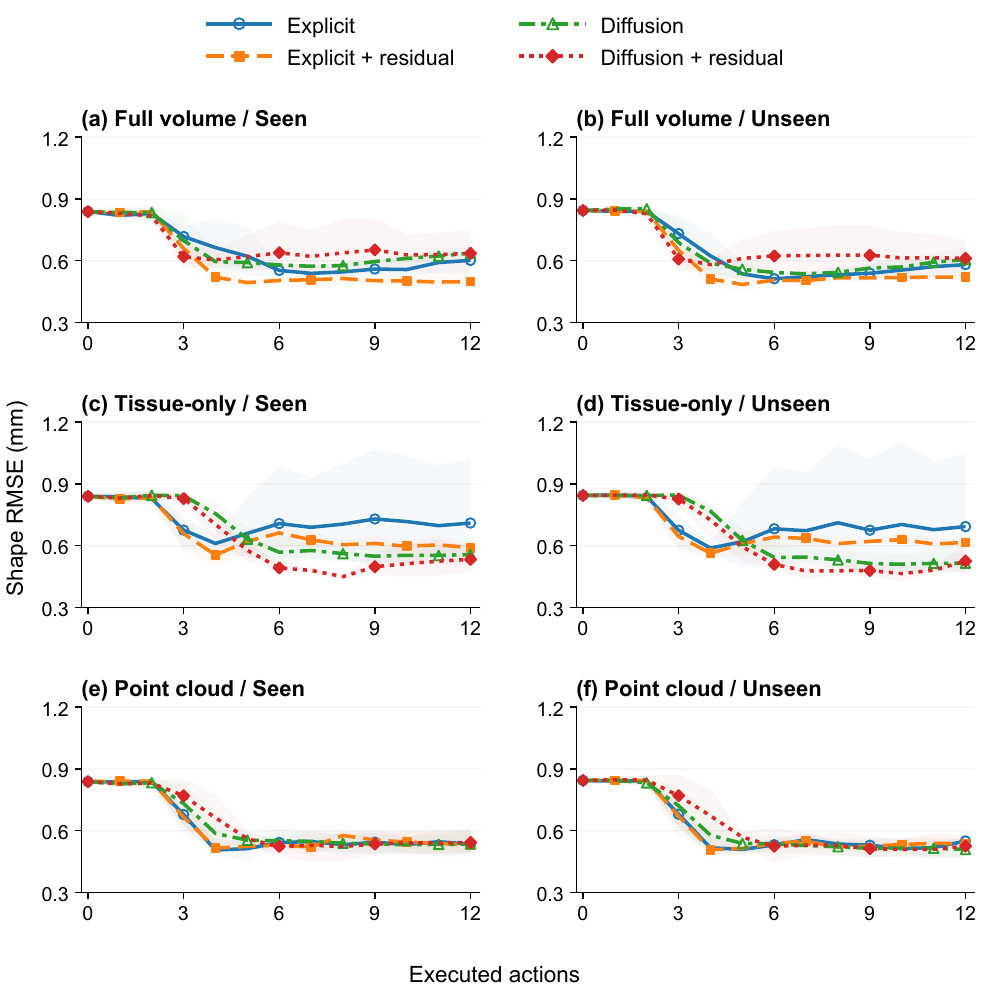}
\caption{\textbf{Phantom depth error across observation steps.} Step 0 is the initial observation; steps 1--12 follow executed actions. Rows show representations and columns show seen/unseen targets. Curves average targets and three sweeps; shading spans sweep means.}
\label{fig:results_depth_curves}
\end{figure}

\begin{figure}[h]
\centering
\includegraphics[width=0.95\linewidth]{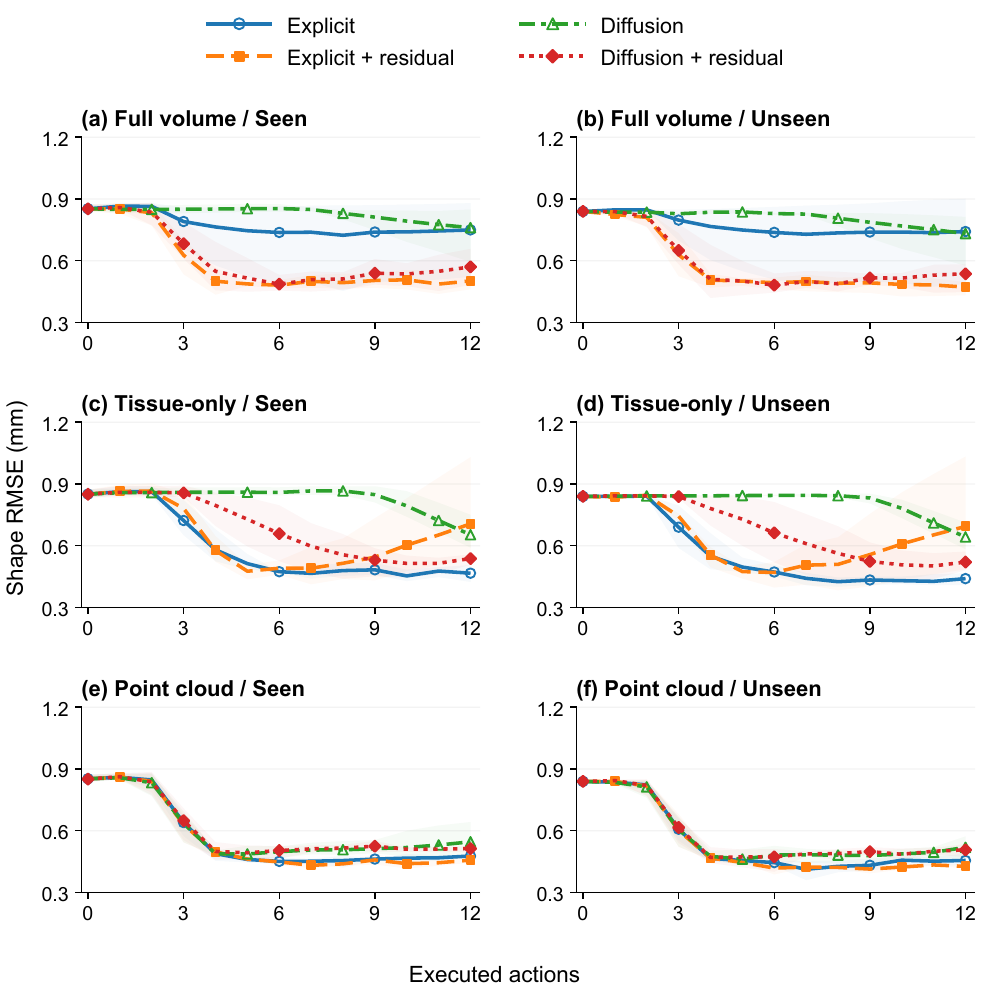}
\caption{
\textbf{Phantom directional error at every observation step.} Aggregation and shading follow Fig.~\ref{fig:results_depth_curves}. Curves show deployments across nine target directions, including four unseen directions.
}
\label{fig:results_direction_curves}
\end{figure}

Figures~\ref{fig:results_depth_curves} and~\ref{fig:results_direction_curves} show Shape RMSE from the initial observation through all twelve executed actions. These curves complement the endpoint results by showing how tissue--target mismatch changes during the rollout.

\subsubsection{Target-dependent action responses}
\label{app:target_dependence}

\paragraph{Target association and response spread.}
We evaluate whether changing the requested target produces a corresponding change in the commanded motion. For each rollout, the twelve local commands are composed in the initial tool frame to obtain a final commanded translation vector and, for the 6-DoF task, a final commanded rotation vector.

For each deployment sweep, the seven depth targets or nine directional targets therefore produce a set of endpoint responses. We characterize this target-dependent response using two complementary quantities. First, distance correlation $r_d$ measures whether targets that are more different also tend to produce more different endpoint commands. Larger $r_d\in[0,1]$ indicates a stronger association between the requested target and the commanded response. Second, response spread $W$ measures how much the endpoint commands actually vary across targets. Larger $W$ indicates greater across-target variation, in millimeters for translation and degrees for rotation.

Specifically, let $\mathbf z_i$ denote the endpoint translation or rotation vector of rollout $i$. Within each sweep, we subtract the average endpoint over all tested targets,
\begin{equation}
\mathbf y_i=\mathbf z_i-\frac{1}{K}\sum_{k=1}^{K}\mathbf z_k,
\end{equation}
where $K=7$ for depth and $K=9$ for direction. Thus, $\mathbf y_i$ measures how the response to target $i$ differs from the typical response in the same sweep.

For target association, depth is represented by its scalar value and direction by $(\cos\phi_i,\sin\phi_i)$. We compare the pairwise distances between target values with the pairwise distances between their corresponding endpoint responses. Thus, $r_d$ is high when target pairs that are farther apart also tend to produce endpoint responses that are farther apart.

Let $A$ and $B$ denote the centered pairwise-distance matrices for the targets and endpoint responses, respectively, following the standard distance-correlation formulation. The target association and response spread are
\begin{equation}
r_d=
\sqrt{
\max\!\left(
0,
\frac{\sum_{ij}A_{ij}B_{ij}}
{\sqrt{\sum_{ij}A_{ij}^{2}\sum_{ij}B_{ij}^{2}}}
\right)
},
\qquad
W=
\sqrt{\frac{1}{n}\sum_i\|\mathbf y_i\|_2^2},
\label{eq:results-dependence}
\end{equation}

Table~\ref{tab:results_target_response} includes all 21 depth and 27 directional rollouts per configuration, including unseen targets.

\begin{table}[t]
\centering
\footnotesize
\setlength{\tabcolsep}{2.5pt}
\renewcommand{\arraystretch}{1.10}
\caption{Target association and across-target response spread for all phantom configurations. $r_d$ measures the association between target differences and endpoint-response differences; larger is stronger. $W$ measures the RMS spread of endpoint responses across targets and is reported in millimeters for translation and degrees for rotation.}
\label{tab:results_target_response}
\begin{tabular*}{\linewidth}{@{\extracolsep{\fill}}llrrrrrr@{}}
\toprule
 & & \multicolumn{2}{c}{3-DoF translation}
 & \multicolumn{2}{c}{6-DoF translation}
 & \multicolumn{2}{c}{6-DoF rotation}\\
\cmidrule(lr){3-4}\cmidrule(lr){5-6}\cmidrule(lr){7-8}
Configuration & $R$
& $r_d$ & $W$
& $r_d$ & $W$
& $r_d$ & $W$\\
\midrule
Full volume / Explicit & -- & 0.577 & 0.368 & 0.421 & 2.174 & 0.461 & 5.940\\
Full volume / Explicit & $\checkmark$ & 0.994 & 1.027 & 0.671 & 2.899 & 0.487 & 6.659\\
Full volume / Diffusion & -- & 0.456 & 0.340 & 0.350 & 0.717 & 0.360 & 0.427\\
Full volume / Diffusion & $\checkmark$ & 0.955 & 0.920 & 0.944 & 5.203 & 0.895 & 11.030\\
\addlinespace[2pt]
Tissue-only / Explicit & -- & 0.774 & 0.385 & 0.810 & 3.328 & 0.894 & 8.999\\
Tissue-only / Explicit & $\checkmark$ & 0.976 & 1.118 & 0.919 & 4.766 & 0.881 & 8.745\\
Tissue-only / Diffusion & -- & 0.808 & 0.149 & 0.603 & 0.063 & 0.698 & 0.121\\
Tissue-only / Diffusion & $\checkmark$ & 0.972 & 1.000 & 0.874 & 6.403 & 0.910 & 12.211\\
\addlinespace[2pt]
Point cloud / Explicit & -- & 0.948 & 1.493 & 0.883 & 3.886 & 0.925 & 8.458\\
Point cloud / Explicit & $\checkmark$ & 0.983 & 1.433 & 0.952 & 4.056 & 0.943 & 8.334\\
Point cloud / Diffusion & -- & 0.924 & 0.495 & 0.917 & 4.554 & 0.896 & 10.740\\
Point cloud / Diffusion & $\checkmark$ & 0.973 & 1.014 & 0.920 & 3.795 & 0.885 & 9.256\\
\bottomrule
\end{tabular*}
\end{table}

\subsubsection{Agreement with demonstrations}
\label{app:demonstration_comparison}

We evaluate how closely each policy rollout matches the demonstrated commands at the same target. For policy configuration $c$, target $k$, and deployment sweep $r$, we sum the command increments over the twelve executed actions:
\begin{equation}
\mathbf u^{\mathrm{tr}}_{c,k,r}
=
\sum_{t=1}^{12}\Delta\mathbf p_{c,k,r,t},
\qquad
\mathbf u^{\mathrm{rot}}_{c,k,r}
=
\sum_{t=1}^{12}\Delta\boldsymbol\theta_{c,k,r,t}.
\end{equation}
These component-wise command sums describe accumulated commanded motion. The target-association analysis (Appendix~\ref{app:target_dependence}) uses endpoints obtained by composing the local motions.

For demonstration $d$ at target $k$, we similarly use its complete recorded sequence of length $L_{k,d}$:
\begin{equation}
\mathbf u^{\mathrm{tr,D}}_{k,d}
=
\sum_{t=1}^{L_{k,d}}\Delta\mathbf p_{k,d,t},
\qquad
\mathbf u^{\mathrm{rot,D}}_{k,d}
=
\sum_{t=1}^{L_{k,d}}\Delta\boldsymbol\theta_{k,d,t}.
\end{equation}
The command-reference analysis uses a balanced subset of ten demonstrations at each of the four phantom depth targets and five phantom directional targets, giving 40 depth and 50 directional reference trajectories. Their average cumulative command is
\begin{equation}
\bar{\mathbf u}^{D}_{k}
=
\frac{1}{10}
\sum_{d=1}^{10}
\mathbf u^{D}_{k,d},
\end{equation}
where translation and rotation are evaluated separately.

For each policy rollout, we measure its distance to (1) the average demonstrated command at the same target and (2) the closest individual demonstration:
\begin{equation}
m^{\mathrm{avg}}_{c,k,r}
=
\left\|
\mathbf u_{c,k,r}
-
\bar{\mathbf u}^{D}_{k}
\right\|_2,
\qquad
m^{\mathrm{near}}_{c,k,r}
=
\min_{1\le d\le10}
\left\|
\mathbf u_{c,k,r}
-
\mathbf u^{D}_{k,d}
\right\|_2.
\label{eq:results-command-reference}
\end{equation}
Lower values indicate closer agreement with the demonstrated commands. The average-reference distance measures agreement with the central demonstrated behavior at that target, while the nearest-demonstration distance measures agreement with the most similar recorded demonstration. Only targets with demonstrations are included: four depth targets and five directional targets. Table~\ref{tab:results_demo_vectors} reports the resulting configuration-level averages. 

\begin{table}[!htbp]
\centering
\footnotesize
\setlength{\tabcolsep}{2.5pt}
\renewcommand{\arraystretch}{1.10}
\caption{Cumulative-command agreement with demonstrations. ``Demo avg.'' is the distance to the average cumulative command of the ten demonstrations at the same target; ``Nearest demo'' is the distance to the closest individual demonstration.}
\label{tab:results_demo_vectors}
\begin{tabular*}{\linewidth}{@{\extracolsep{\fill}}llrrrrrr@{}}
\toprule
 & &
\multicolumn{2}{c}{3-DoF translation} &
\multicolumn{2}{c}{6-DoF translation} &
\multicolumn{2}{c}{6-DoF rotation}\\
\cmidrule(lr){3-4}
\cmidrule(lr){5-6}
\cmidrule(lr){7-8}
Configuration & $R$
& \shortstack{Demo\\avg.}
& \shortstack{Nearest\\demo}
& \shortstack{Demo\\avg.}
& \shortstack{Nearest\\demo}
& \shortstack{Demo\\avg.}
& \shortstack{Nearest\\demo}\\
\midrule
Full volume / Explicit & -- & 2.032 & 1.366 & 7.551 & 6.743 & 14.403 & 10.959\\
Full volume / Explicit & $\checkmark$ & 0.745 & 0.726 & 3.200 & 2.114 & 9.174 & 6.643\\
Full volume / Diffusion & -- & 2.297 & 1.867 & 6.330 & 5.426 & 11.056 & 6.931\\
Full volume / Diffusion & $\checkmark$ & 1.840 & 1.152 & 2.566 & 1.859 & 4.664 & 3.037\\
\addlinespace[2pt]
Tissue-only / Explicit & -- & 2.213 & 1.958 & 3.328 & 2.337 & 7.096 & 5.443\\
Tissue-only / Explicit & $\checkmark$ & 1.346 & 0.895 & 4.672 & 3.859 & 8.010 & 6.070\\
Tissue-only / Diffusion & -- & 3.248 & 2.357 & 5.989 & 4.999 & 10.958 & 6.906\\
Tissue-only / Diffusion & $\checkmark$ & 2.168 & 1.372 & 3.727 & 2.935 & 9.821 & 7.943\\
\addlinespace[2pt]
Point cloud / Explicit & -- & 0.840 & 0.637 & 2.255 & 1.714 & 5.850 & 4.599\\
Point cloud / Explicit & $\checkmark$ & 0.732 & 0.567 & 1.604 & 1.247 & 4.535 & 3.413\\
Point cloud / Diffusion & -- & 1.881 & 1.179 & 3.566 & 2.872 & 9.619 & 7.677\\
Point cloud / Diffusion & $\checkmark$ & 1.646 & 0.954 & 3.686 & 2.871 & 6.686 & 5.216\\
\bottomrule
\end{tabular*}
\end{table}

\subsubsection{Step-by-step action trajectories}
\label{app:command_variation}
\label{app:six_dof_components}

\begin{figure*}[!htbp]
\centering
\includegraphics[width=\textwidth]{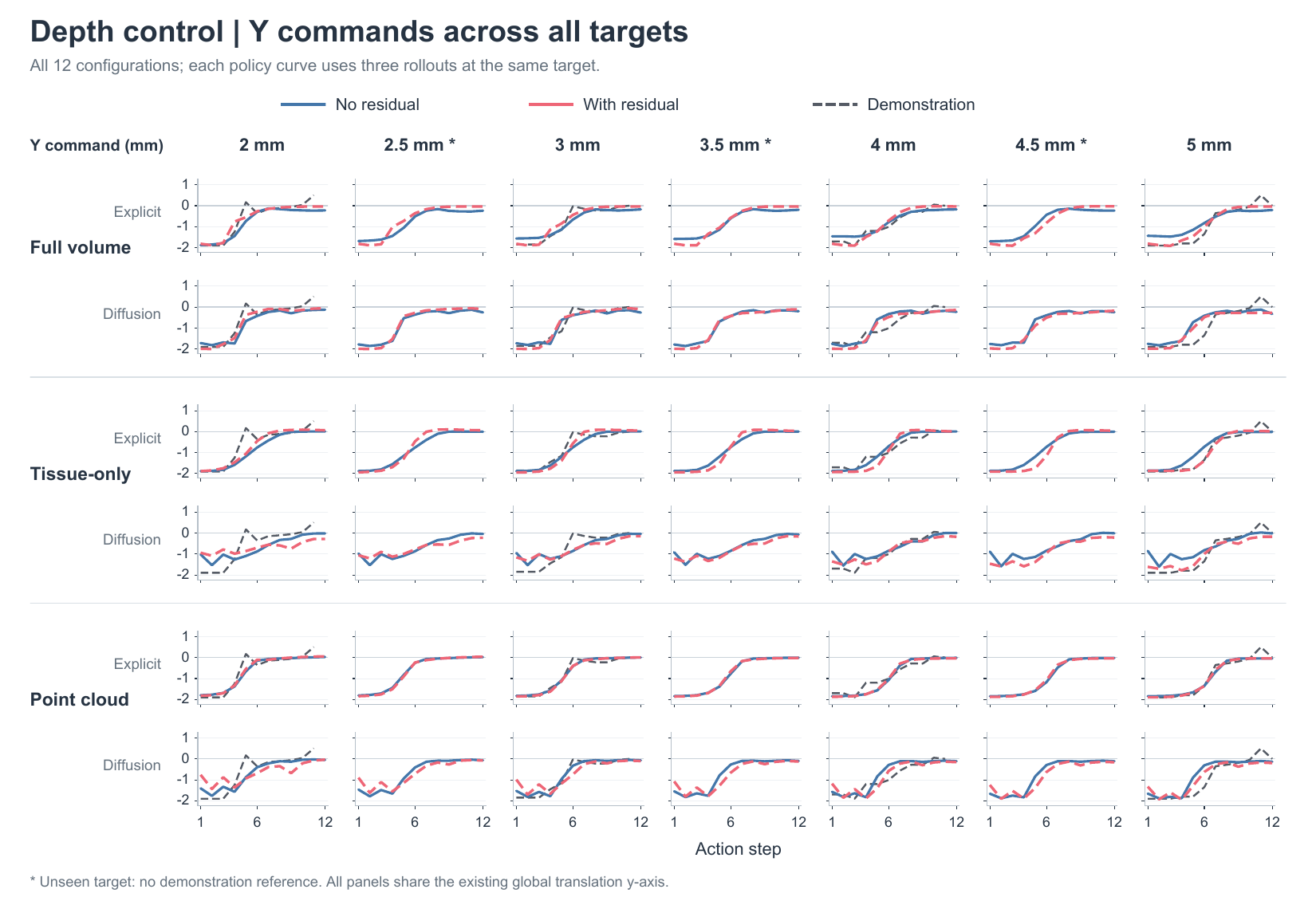}
\caption{\textbf{Depth-task $y$ translation commands across all targets.} Columns show the seven depth targets, with asterisks marking unseen intermediate targets. Rows show all twelve policy configurations grouped by representation and action model. }
\label{fig:actions_depth_y_all_targets}
\end{figure*}

\begin{figure*}[!htbp]
\centering
\includegraphics[width=\textwidth]{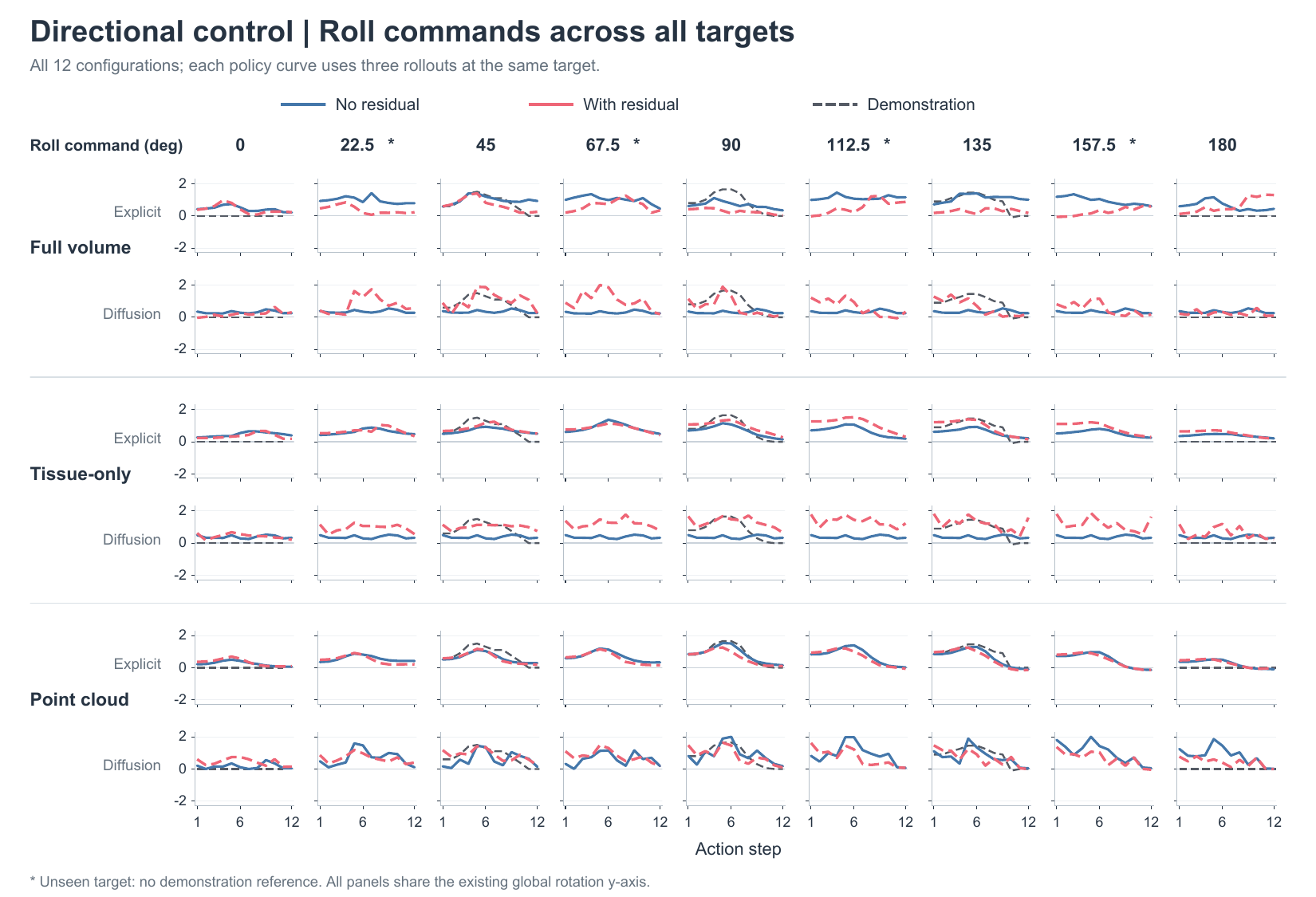}
\caption{\textbf{Directional-task roll commands across all targets.} Columns show the nine tested target directions, with asterisks marking unseen intermediate targets. Rows: twelve policy configurations.}
\label{fig:actions_direction_roll_all_targets}
\end{figure*}

\begin{figure*}[!htbp]
\centering
\includegraphics[width=\textwidth]{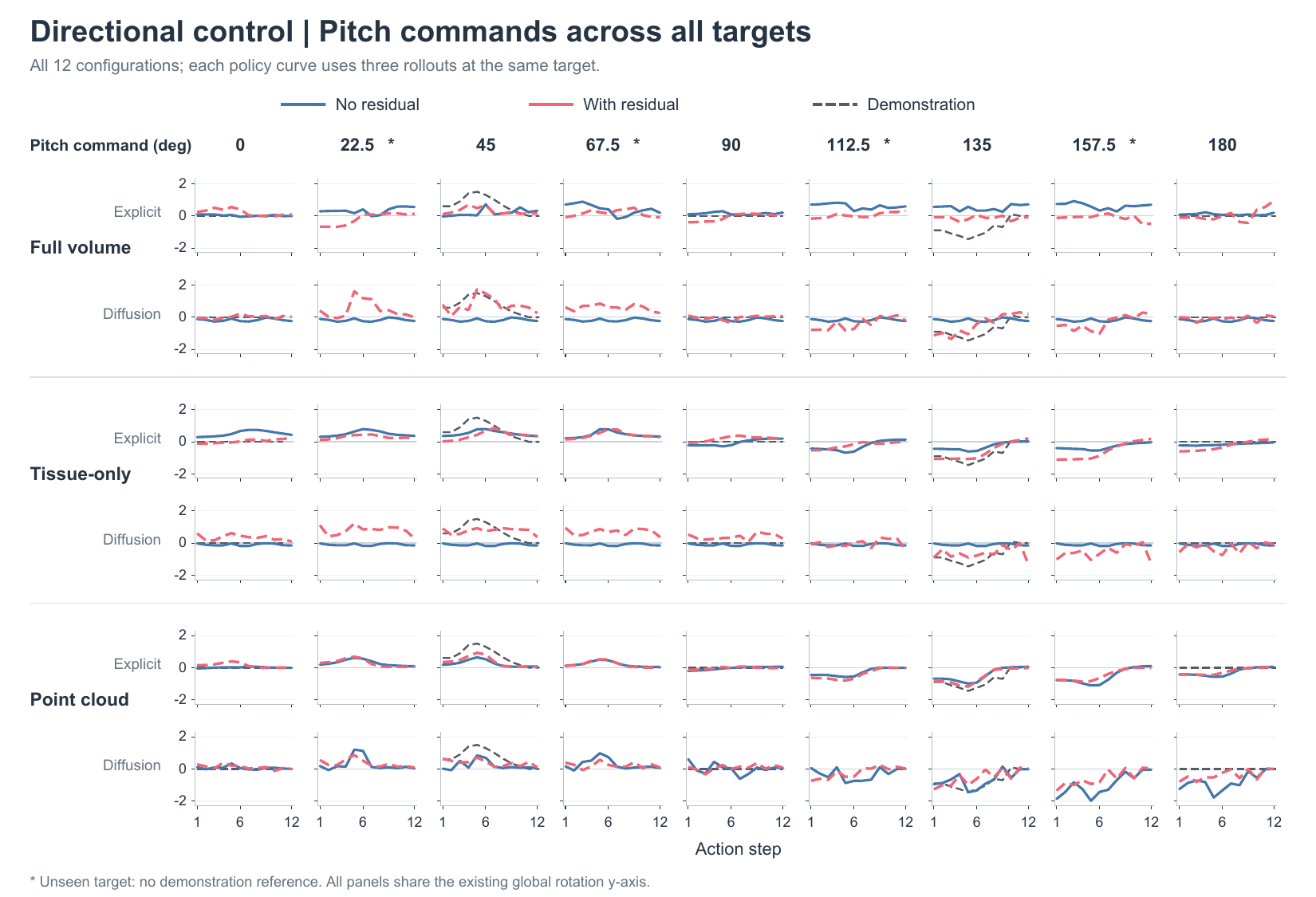}
\caption{\textbf{Directional-task pitch commands across all targets.}}
\label{fig:actions_direction_pitch_all_targets}
\end{figure*}

\begin{figure*}[t]
\centering
\includegraphics[width=\textwidth]{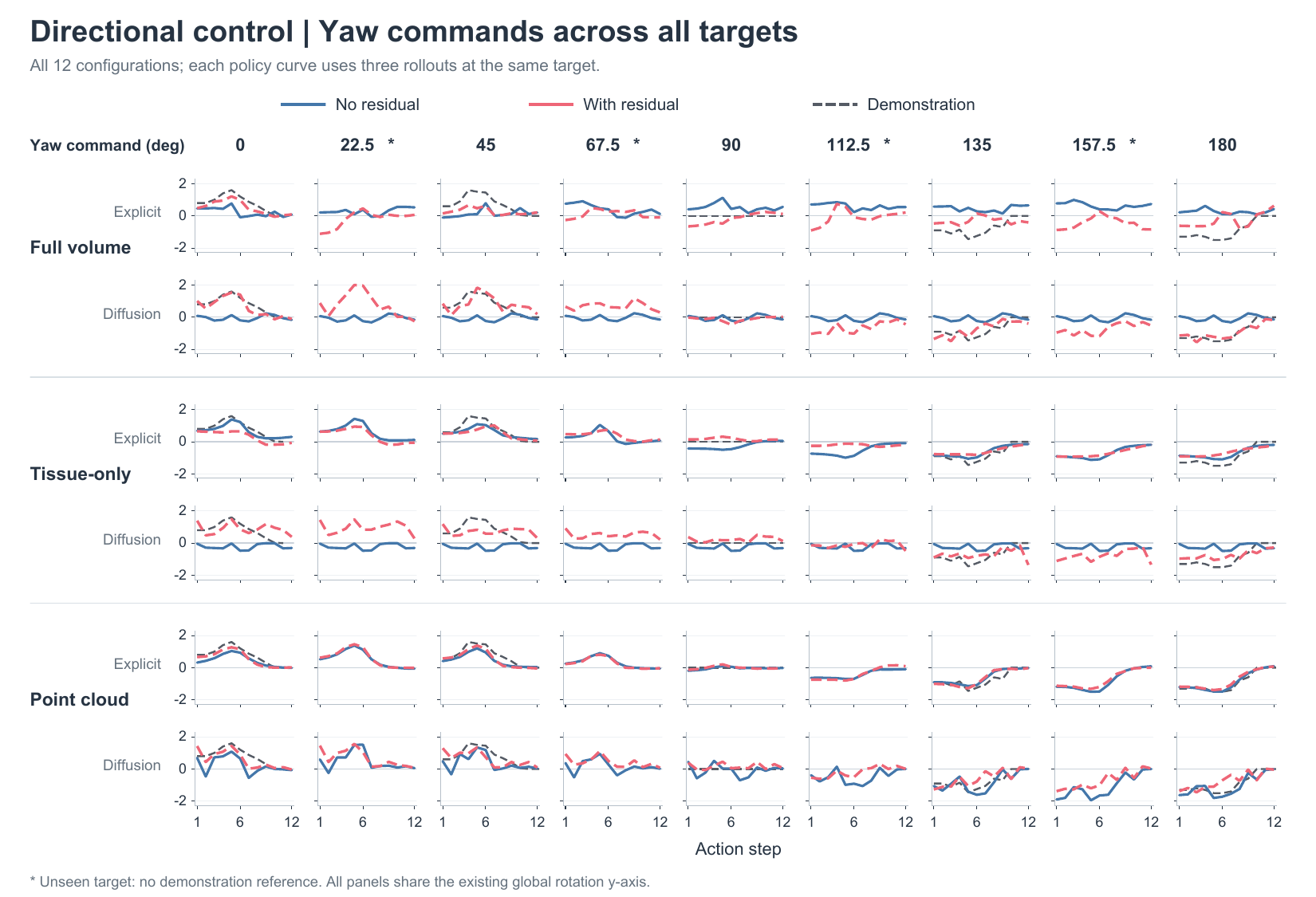}
\caption{\textbf{Directional-task yaw commands across all targets.} }
\label{fig:actions_direction_yaw_all_targets}
\end{figure*}

Figures~\ref{fig:actions_depth_y_all_targets}--\ref{fig:actions_direction_yaw_all_targets} summarize the step-by-step action trajectories using one command component per figure. For the depth task, we report the $y$ translation command because it is the primary deformation axis. For the directional task, we report the three rotation components, roll, pitch, and yaw, to show how rotational control changes across target directions.

Within each figure, columns enumerate all tested targets and rows enumerate the twelve policy configurations grouped by representation and action model. Each policy curve is the mean of the three deployment sweeps at that target. At demonstrated targets, the gray reference curve is the mean demonstration command at each action index. 

\subsection{\textit{Ex vivo} porcine results}
\label{app:exvivo}
\label{app:exvivo-results}

We evaluate Full-volume Diffusion and Voxelized point-cloud Explicit, both with residual conditioning. Each configuration is tested at five depth and five directional targets with three deployments per target, yielding 60 twelve-action rollouts. Target conditions are listed in Table~\ref{tab:evaluation_target_conditions}; geometric summaries follow Eq.~\eqref{eq:results-aggregation} with $K=5$ for each task.

\subsubsection{Shape agreement}

\textbf{Endpoint results:}
Table~\ref{tab:exvivo-shape} reports initial and final Shape RMSE and HD95 for both tasks, while Table~\ref{tab:exvivo-directed} separates the two distance directions to relate endpoint errors to the observed deformation.

\begin{table}[!htbp]
\centering
\small
\setlength{\tabcolsep}{3pt}
\renewcommand{\arraystretch}{1.10}
\caption{Ex vivo endpoint geometry (mm) for 60 rollouts, with 15 per row. Both configurations use residual conditioning. $D_0$ and $D_{12}$ are means $\pm$ sample SD across three target-averaged sweeps (Eq.~\eqref{eq:results-aggregation}); HD95 columns report means.}
\label{tab:exvivo-shape}
\begin{tabular*}{\linewidth}{@{\extracolsep{\fill}}lrrrrr@{}}
\toprule
Configuration & $D_0$ & $D_{12}$ & $H_{95,0}$ & $H_{95,12}$ & Improved \\
\midrule
\multicolumn{6}{l}{\textit{3-DoF depth}} \\
Full volume / Diffusion & $0.650\pm0.044$ & $1.102\pm0.045$ & 2.003 & 4.390 & 0/15 \\
Point cloud / Explicit & $0.592\pm0.035$ & $1.220\pm0.088$ & 1.906 & 4.690 & 0/15 \\
\midrule
\multicolumn{6}{l}{\textit{6-DoF directional}} \\
Full volume / Diffusion & $0.985\pm0.021$ & $0.961\pm0.024$ & 3.376 & 2.892 & 8/15 \\
Point cloud / Explicit & $1.022\pm0.028$ & $1.081\pm0.038$ & 3.434 & 2.978 & 4/15 \\
\bottomrule
\end{tabular*}
\end{table}

\textbf{Two directions of geometric mismatch:}
We separate the target-to-observation and observation-to-target RMSE terms in Eq.~\eqref{eq:results-shape-rmse}, together with the directed q95 terms used in HD95 (Table~\ref{tab:exvivo-directed}). Target-to-observation RMSE decreases from the initial to final state in 58 of 60 rollouts; observation-to-target RMSE increases in all 60. The two exceptions are Point-cloud depth rollouts at the 2\,mm target. For q95, the corresponding decrease and increase occur in all 60 rollouts.

\begin{table}[t]
\centering
\footnotesize
\setlength{\tabcolsep}{3pt}
\renewcommand{\arraystretch}{1.10}
\caption{Directed components of Shape RMSE and HD95. Each entry shows the initial-to-final mean in millimeters. RMSE refers to the directed terms in Eq.~\eqref{eq:results-shape-rmse}; q95 refers to the two percentiles whose maximum defines HD95.}
\label{tab:exvivo-directed}
\begin{tabular*}{\linewidth}{@{\extracolsep{\fill}}lrrrr@{}}
\toprule
& \multicolumn{2}{c}{Target $\to$ observation}
& \multicolumn{2}{c}{Observation $\to$ target} \\
\cmidrule(lr){2-3}\cmidrule(lr){4-5}
Configuration & RMSE & q95 & RMSE & q95 \\
\midrule
\multicolumn{5}{l}{\textit{3-DoF depth}} \\
Human reference & $0.760\to0.395$ & $1.988\to0.759$ & $0.486\to1.544$ & $1.216\to4.344$ \\
Full volume / Diffusion & $0.767\to0.334$ & $2.003\to0.655$ & $0.499\to1.520$ & $1.242\to4.390$ \\
Point cloud / Explicit & $0.732\to0.530$ & $1.906\to0.951$ & $0.394\to1.639$ & $0.972\to4.690$ \\
\midrule
\multicolumn{5}{l}{\textit{6-DoF directional}} \\
Human reference & $1.268\to0.484$ & $3.308\to0.936$ & $0.523\to1.169$ & $1.248\to2.760$ \\
Full volume / Diffusion & $1.302\to0.499$ & $3.376\to0.983$ & $0.488\to1.259$ & $1.161\to2.892$ \\
Point cloud / Explicit & $1.321\to0.659$ & $3.434\to1.192$ & $0.579\to1.376$ & $1.398\to2.978$ \\
\bottomrule
\end{tabular*}
\end{table}

\begin{figure}[t]
\centering
\includegraphics[width=0.95\linewidth]{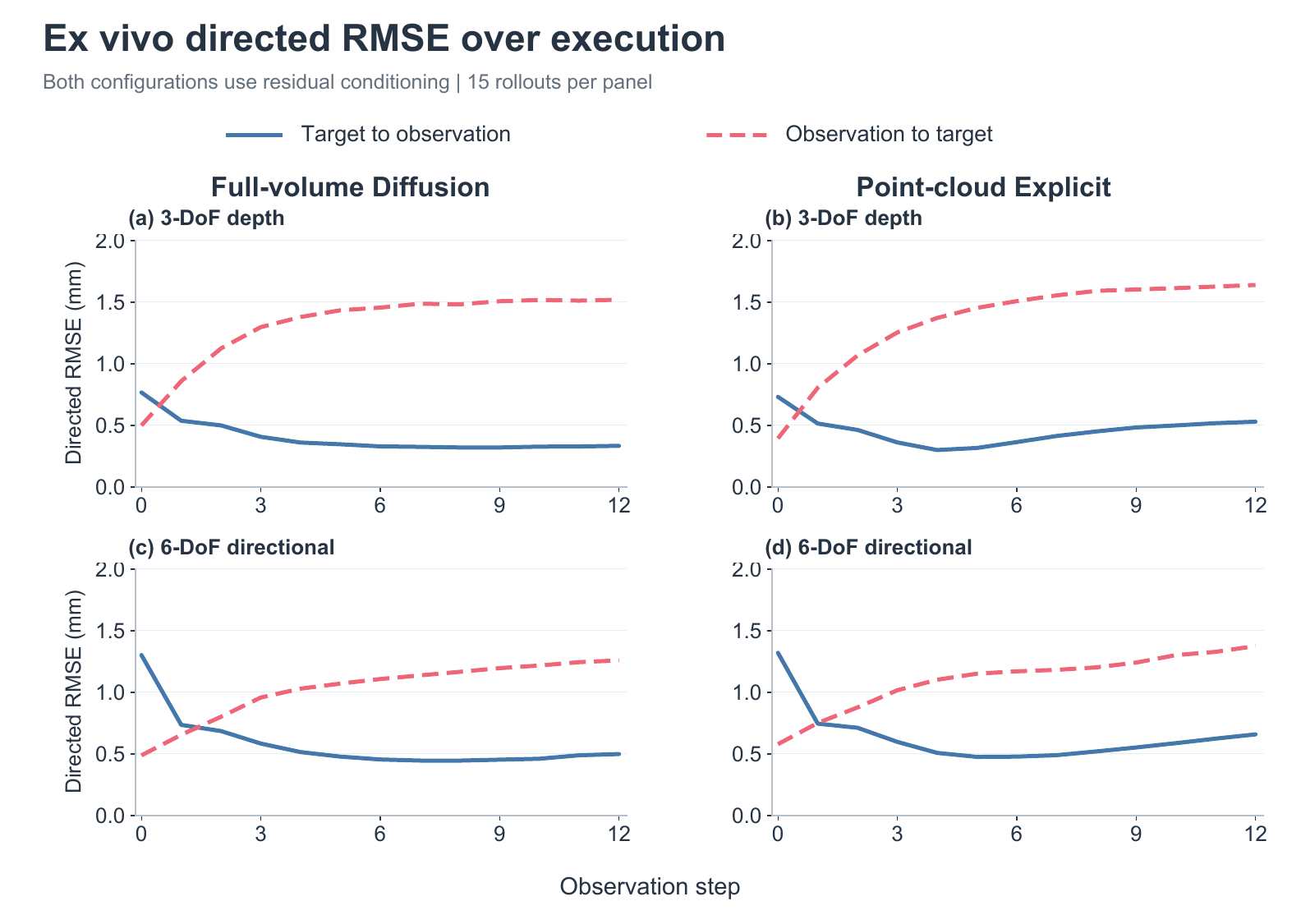}
\caption{\textbf{Directed \textit{ex vivo} RMSE over execution.}
Rows show depth and directional tasks, and columns show the two policy configurations. Blue denotes target-to-observation RMSE and coral denotes observation-to-target RMSE.}
\label{fig:exvivo-directed-rmse}
\end{figure}

Figure~\ref{fig:exvivo-directed-rmse} shows the two RMSE terms over execution. Target-to-observation error falls during the early actions and rises slightly again later, particularly for Point-cloud Explicit. Observation-to-target error generally increases throughout execution.

\paragraph{Relation to tissue deformation.}
The initial decrease in target-to-observation RMSE is consistent with observed tissue approaching the desired depression. Its later rebound may reflect tissue moving away from some aligned target regions. For example, continued indentation could improve alignment at the center while widening the depression and increasing mismatch along its sidewalls.

Observation-to-target RMSE also includes observed tissue outside the target support. If these regions move farther from the target or occupy a larger fraction of the observed support, their contribution can outweigh the improvement at the center and increase this error. These trends are therefore consistent with partial alignment toward the intended indentation, while the width, sidewalls, and surrounding tissue remain mismatched. 

\subsubsection{Action responses}
\label{app:exvivo-action}

\textbf{Target dependence:}
We next evaluate whether the executed commands vary systematically with the requested target, using the same target-association score $r_d$ and response spread $W$ as in the phantom study. Table~\ref{tab:exvivo_target_response} summarizes all retained ex vivo rollouts. 

\begin{table}[t]
\centering
\footnotesize
\setlength{\tabcolsep}{4pt}
\renewcommand{\arraystretch}{1.10}
\caption{Target-dependent ex vivo command responses. Statistics use the same definitions as in the phantom action-response analysis. $r_d$ is the target-association score and $W$ is the response spread.}
\label{tab:exvivo_target_response}
\begin{tabular*}{\linewidth}{@{\extracolsep{\fill}}lllrr@{}}
\toprule
Task & Configuration & Response & $r_d$ & $W$ \\
\midrule
\multicolumn{5}{l}{\textit{3-DoF depth}} \\
Depth & Full volume / Diffusion & Translation & 0.854 & 1.221 \\
Depth & Point cloud / Explicit & Translation & 0.660 & 0.692 \\
\midrule
\multicolumn{5}{l}{\textit{6-DoF directional}} \\
Directional & Full volume / Diffusion & Translation & 0.926 & 0.984 \\
Directional & Full volume / Diffusion & Rotation & 0.974 & 2.051 \\
Directional & Point cloud / Explicit & Translation & 0.949 & 0.972 \\
Directional & Point cloud / Explicit & Rotation & 0.968 & 2.051 \\
\bottomrule
\end{tabular*}
\end{table}

\textbf{Agreement with demonstrations:}
Table~\ref{tab:exvivo_demo_agreement} reports component-wise cumulative-command agreement using Eq.~\eqref{eq:results-command-reference}, evaluated at demonstrated depths ($2$, $3$, $4$\,mm) and directions ($0^\circ$, $45^\circ$, $90^\circ$).

\begin{table}[t]
\centering
\footnotesize
\setlength{\tabcolsep}{3.6pt}
\renewcommand{\arraystretch}{1.10}
\caption{Ex vivo cumulative-command agreement with demonstrations.
``Demo avg.'' and ``Nearest demo'' use the same component-wise command
sums and reference distances as Table~\ref{tab:results_demo_vectors}.}
\label{tab:exvivo_demo_agreement}
\begin{tabular*}{\linewidth}{@{\extracolsep{\fill}}lrrrrrr@{}}
\toprule
& \multicolumn{2}{c}{3-DoF translation}
& \multicolumn{2}{c}{6-DoF translation}
& \multicolumn{2}{c}{6-DoF rotation} \\
\cmidrule(lr){2-3}
\cmidrule(lr){4-5}
\cmidrule(lr){6-7}
Configuration
& \shortstack{Demo\\avg.}
& \shortstack{Nearest\\demo}
& \shortstack{Demo\\avg.}
& \shortstack{Nearest\\demo}
& \shortstack{Demo\\avg.}
& \shortstack{Nearest\\demo} \\
\midrule
Full volume / Diffusion
& 1.148 & 0.697
& 0.790 & 0.852
& 0.911 & 0.478 \\
Point cloud / Explicit
& 0.717 & 0.586
& 1.315 & 0.914
& 1.526 & 1.069 \\
\bottomrule
\end{tabular*}
\end{table}

\subsubsection{Ex vivo OCT and Color Image examples}
\label{app:exvivo-qualitative}

\begin{figure}[!htbp]
\centering
\includegraphics[width=\linewidth]{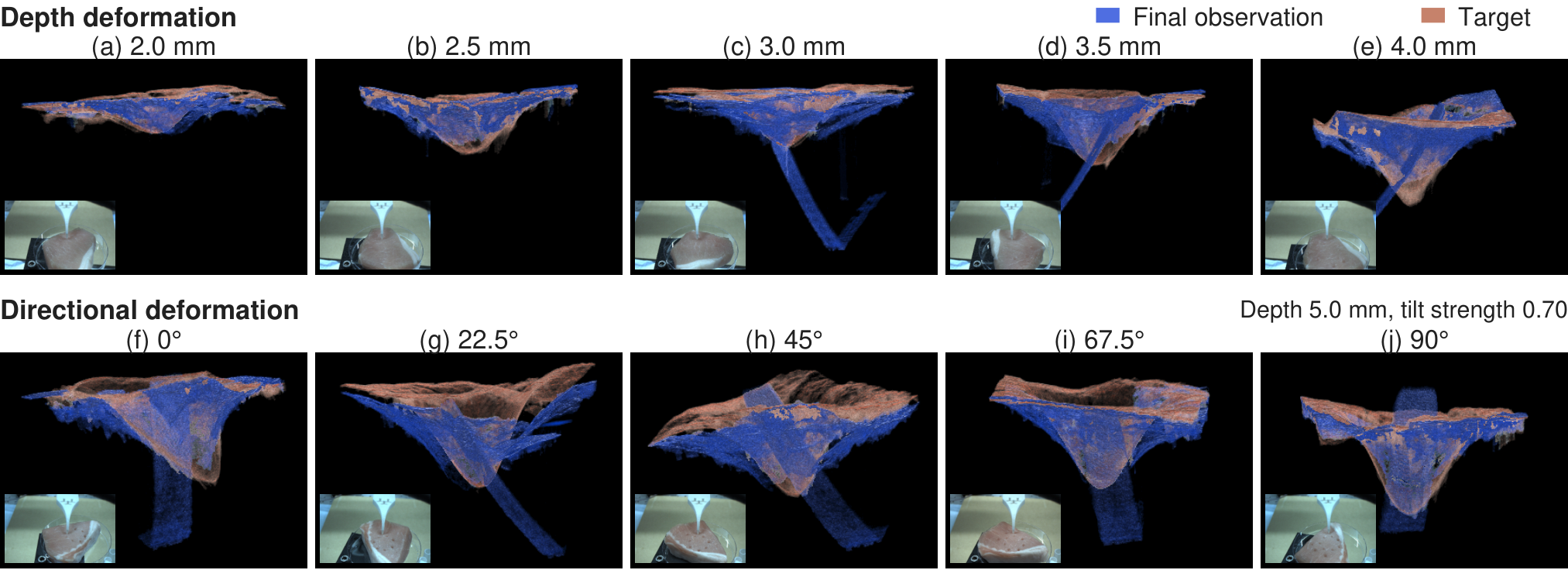}
\caption{\textbf{Representative ex vivo endpoint comparisons for Full-volume Diffusion + residual.} Top row: depth targets $2.0$--$4.0$\,mm. Bottom row: directional targets $0^\circ$--$90^\circ$ at depth $5.0$\,mm and tilt strength $0.70$. Blue shows the final observation and brown shows the target.}
\label{fig:exvivo_full_examples}
\end{figure}

\begin{figure}[h]
\centering
\includegraphics[width=\linewidth]{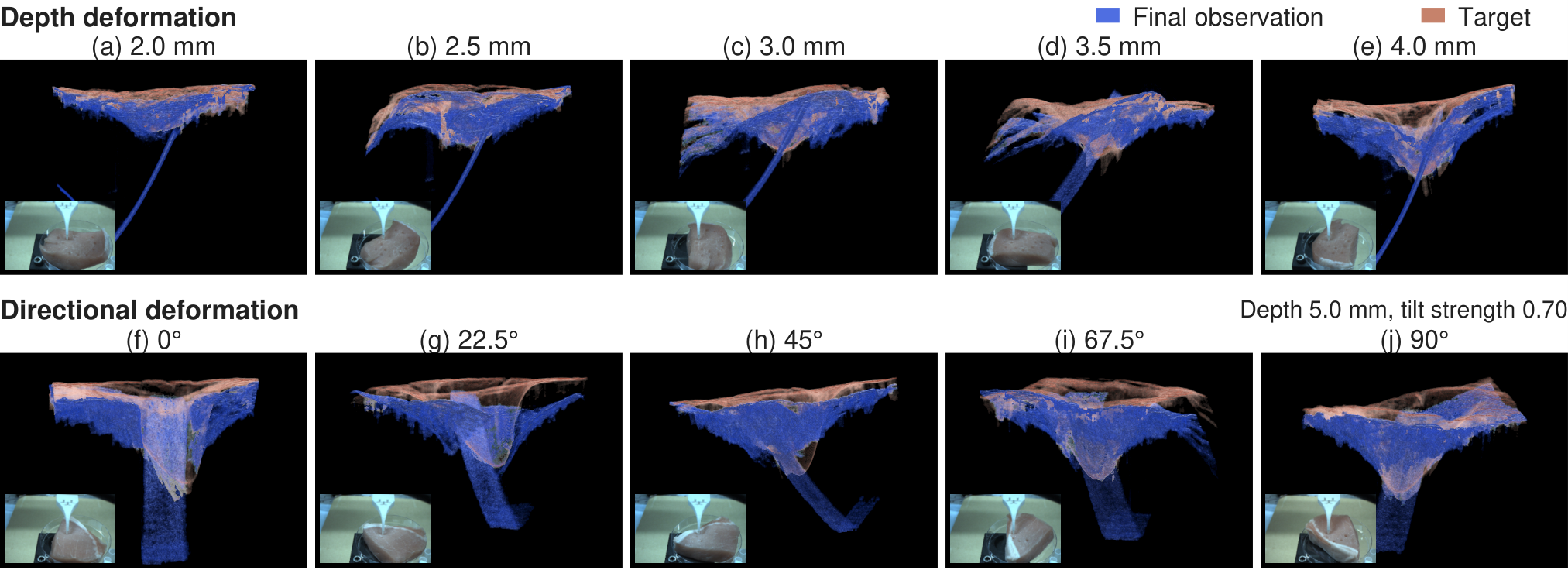}
\caption{\textbf{Representative ex vivo endpoint comparisons for Voxelized point-cloud Explicit with residual conditioning.}}
\label{fig:exvivo_point_examples}
\end{figure}

Figures~\ref{fig:exvivo_full_examples} and~\ref{fig:exvivo_point_examples} compare the final observed and target surfaces for both configurations in scanner coordinates. RGB insets show the corresponding tool--tissue interactions.

\section{Operator-Guided Demonstration Data Collection}
\label{user_data}

Two operators independently collected demonstrations through an interactive interface, selecting incremental 6-DoF tooltip motions from the current tissue state and target morphology without prescribed trajectories. Each trajectory records an initial OCT scan and tooltip pose followed by 9--15 actions and the resulting scans and poses. Recorded trajectories are reviewed using tissue--target overlays, retaining demonstrations with feasible interactions and progressive target alignment for training.

\begin{figure}[!htbp]
\centering
\includegraphics[width=0.85\linewidth]{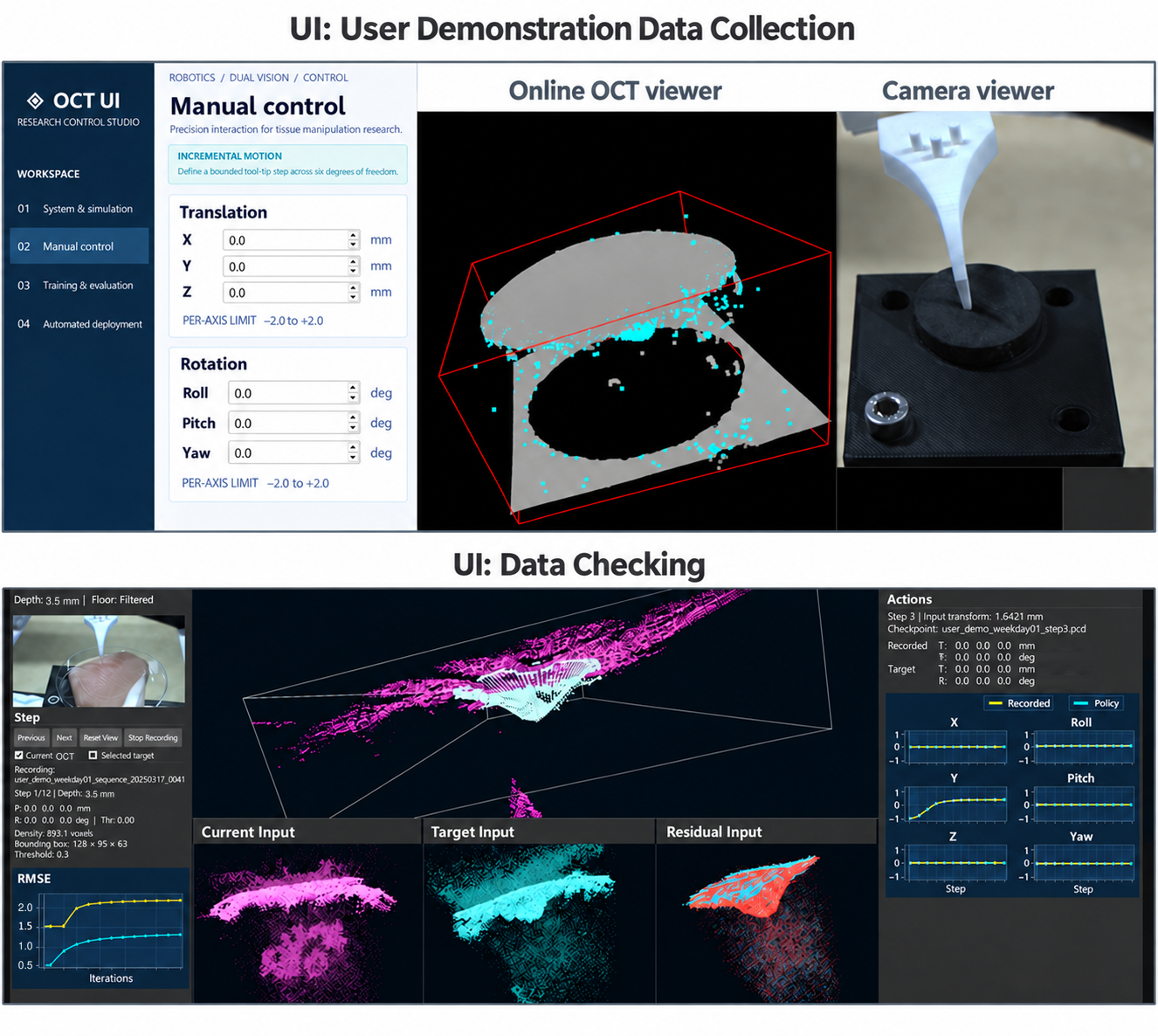}
\caption{\textbf{Demonstration collection and verification.} Top: tooltip control with OCT and RGB feedback. Bottom: trajectory review with tissue--target overlays and geometric error.}
\label{fig:data_checking}
\end{figure}

\end{document}